\documentclass[journal]{IEEEtran}
\usepackage{amsmath,amsfonts,amssymb}
\usepackage{algorithm}
\usepackage{algorithmicx}
\usepackage{algpseudocode}
\algrenewcommand\alglinenumber[1]{#1}
\algrenewtext{EndIf}{\textbf{end}}
\usepackage{array}
\usepackage{textcomp}
\usepackage{stfloats}
\usepackage{verbatim}
\usepackage{graphicx}
\usepackage{cite}
\usepackage{multirow}
\usepackage[hidelinks]{hyperref}
\usepackage{cleveref}
\usepackage{xcolor}
\usepackage{pifont}
\usepackage{makecell}
\usepackage{xspace}
\usepackage{todonotes}
\usepackage{subcaption}
\def\BibTeX{{\rm B\kern-.05em{\sc i\kern-.025em b}\kern-.08em
    T\kern-.1667em\lower.7ex\hbox{E}\kern-.125emX}}
\usepackage{balance}
\usepackage{siunitx}
\usepackage{booktabs}
\usepackage{threeparttable}
\newcommand{\mathset}[1]{\mathcal{#1}}
\newcommand{\mathvec}[1]{\mathbf{#1}}

\newcommand{\cmark}{\textcolor{green!60!black}{\checkmark}} 
\newcommand{\xmark}{\textcolor{red}{\ding{55}}}

\usepackage{soul}

\DeclareMathOperator{\Conn}{Conn}
\DeclareMathOperator{\Class}{Class}
\DeclareMathOperator{\Cost}{Cost}
\DeclareMathOperator{\Heur}{Heur}

\begin{document}
\title{SE(2) Navigation Mesh}
\author{
\thanks{}}

\author{Shuyang Shi, Kaixian Qu, Changan Chen, Ines Kast, Yuntao Ma, Marco Hutter
\thanks{Robotic Systems Lab, ETH Zurich, Switzerland. e-mail: kaixqu@ethz.ch. Corresponding author: Kaixian Qu.}%
}

\markboth{}%
{}

\maketitle

\begin{abstract}

Global navigation for ground robots in complex multi-level environments requires representations that accurately capture traversable regions while enabling efficient path planning. Current approaches present key limitations: Point clouds and volumetric occupancy maps lack explicit surface structure for traversability estimation, whereas direct pathfinding on dense triangle meshes is computationally prohibitive. Navigation meshes mitigate these challenges through polygonal abstraction of the underlying mesh, but assume yaw-invariant traversability, rendering them unsuitable for non-circular robots in constrained spaces. We propose SE(2) Navigation Mesh (SE(2) NavMesh), a polygonal representation of traversable regions that encodes yaw-dependent traversability. Our method evaluates traversability using footprint masks and constructs a graph over yaw-specific layers with explicit translational and rotational connectivity. Grounded in this representation, we develop an A*-String Pulling-A* (ASA) pathfinding strategy that hierarchically optimizes robot position and heading. We also present an online method that incrementally updates the SE(2) NavMesh from streaming point clouds during concurrent geometry reconstruction. In simulation, the SE(2) NavMesh captures over \SI{50}{\percent} more traversable area than classical NavMeshes, and the SE(2) NavMesh + ASA pipeline consistently outperforms sampling-based baselines in constrained environments. Extensive real-world experiments on a physical robot validate real-time online generation and successful navigation across multiple environments.

\end{abstract}


\section{Introduction}

Ground robots are increasingly deployed for infrastructure and industrial facility inspection~\cite{lattanzi2017review, halder2023robots}, owing to their ability to carry heavy payloads and support long-duration missions. Traditionally, these operations have been performed by wheeled robots on relatively flat terrain~\cite{miura2020plant, kim2019uav, fischer2024evaluation}. However, many facilities---such as chemical plants, power stations, and oil rigs---contain stairs, elevated platforms, and uneven terrain that are inaccessible to wheeled robots. Recent advances in legged locomotion have enabled legged robots to traverse rough terrain, stairs, and narrow structures~\cite{miki2022wild, agarwal2023legged}, making them promising platforms for inspection tasks~\cite{kolvenbach2020towards, afsari2021fundamentals}. 

\begin{figure}[t]
    \centering     
    \includegraphics[width=0.98\linewidth]{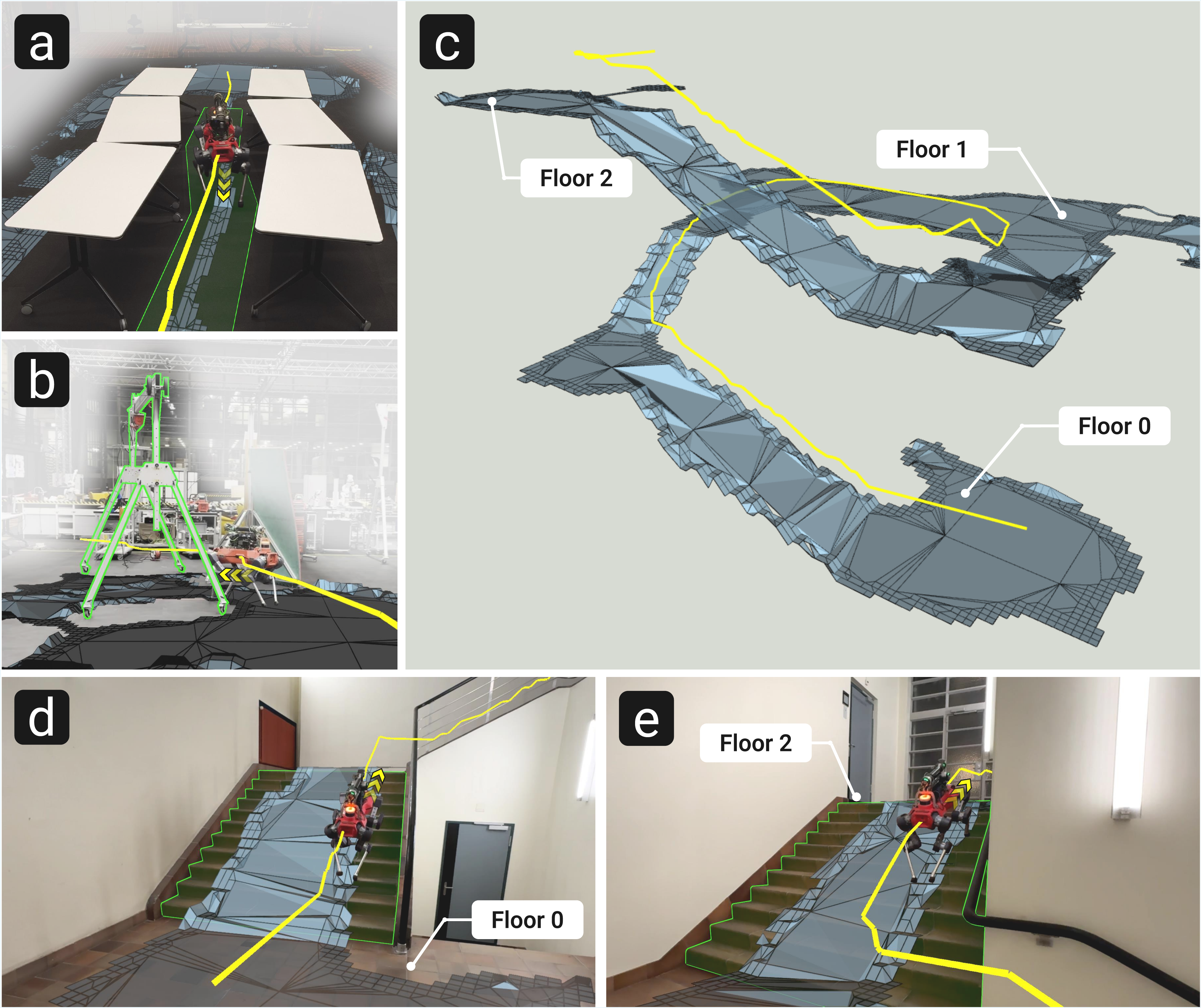}    
    \caption{Real-world robot navigation on the SE(2) NavMesh. Yellow polylines indicate planned paths, and blue polygons represent traversable regions. SE(2) NavMesh captures yaw-dependent traversability in constrained scenes such as (a) narrow passages (\SI{0.8}{\meter} wide in this example, only \SI{0.27}{\meter} wider than the robot) and (b) overhanging obstacles. It also supports multi-level structures (c), enabling successful stair traversal across floors as demonstrated in (d) and (e).}
    \label{fig:se2_navmesh}
\end{figure}

Despite these advancing locomotion capabilities, accurately representing traversable regions that capture complex geometry while enabling efficient global navigation remains an open challenge. Point clouds are a common representation of complex environments~\cite{perez20193d, yang2022far, kong2023marsim}.
Their lack of structural organization, however, makes point clouds inefficient for traversability estimation, as geometric surfaces and connectivity must be inferred through additional processing~\cite{liu2015robotic, krusi2017driving}, leading to substantial computational overhead in large-scale environments~\cite{Yang2025tomography}. From point cloud observations, volumetric representations such as occupancy maps, truncated signed distance fields (TSDFs), and Euclidean signed distance fields (ESDFs) can be constructed~\cite{oleynikova2017voxblox, hornung2013octomap}. These representations enable efficient 3D obstacle reasoning and are widely used for aerial robot navigation~\cite{zhou2020ego, ren2024rog}. Nevertheless, they do not explicitly preserve terrain surface structure, making fine-grained traversability estimation difficult without resorting to point cloud-based geometric analysis or learned models~\cite{wang2023towards,frey2022locomotion}. At a lower dimensionality, 2D grid maps~\cite{elfes2002using} and elevation maps~\cite{fankhauser2014robot, fankhauser2018probabilistic} discretize the environment into regular grids. 2D grid maps are effective in simple planar environments~\cite{fischer2024evaluation, kim2019uav} but fail to represent uneven terrain or multi-floor scenes. Elevation maps, while widely used for legged robot locomotion and local navigation~\cite{fankhauser2014robot, fankhauser2018probabilistic, miki2022elevation}, are inherently single-layer and cannot represent multi-level structures where surfaces overlap at the same planar location.

Moving to surface-based representations, triangle meshes are commonly used in large-scale 3D scene datasets such as HM3D~\cite{ramakrishnan2021habitat} and HSSD~\cite{khanna2023hssd}.
By preserving detailed surface geometry and adjacency relationships between surfaces, triangle meshes support traversability analysis and graph-based pathfinding~\cite{ruetz2019ovpc, puetz2021trianglemesh}. Yet, accurately representing complex environments requires dense triangle meshes, which incur substantial pathfinding overhead. Navigation Meshes (NavMeshes) address this by abstracting traversable regions into connected convex polygons~\cite{van2016comparative} for pathfinding and are widely used in game engines~\cite{unity, unreal} and, more recently, in robot navigation~\cite{habitatchallenge2023,brandao2020gaitmesh}. However, classical NavMesh~\cite{mononen2014recastnav} assumes yaw-invariant traversability. While this simplifies mesh construction and planning, it fails to capture yaw-dependent traversability for ground robots with non-circular footprints in narrow environments~\cite{brandao2020gaitmesh}. This limitation motivates a richer representation that explicitly accounts for robot heading during both navigation mesh construction and path planning.

\begin{figure*}[t]
    \centering
    \includegraphics[width=0.95\textwidth]{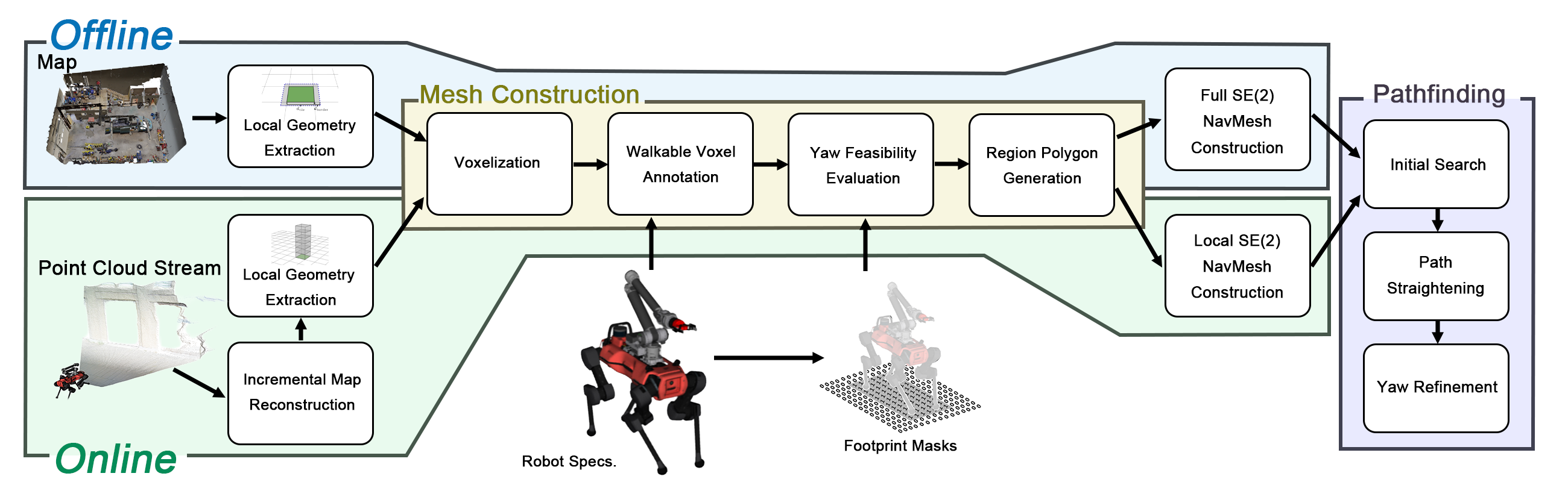}
    \caption{Overview of the SE(2) NavMesh system. The offline pipeline constructs the mesh from a pre-built map, whereas the online pipeline updates it from a point cloud stream.  During mesh construction, map geometry is processed using the robot-specific traversal constraints including traversal capabilities, height, and footprint masks. Path planning is then performed on the generated SE(2) NavMesh using a three-stage pathfinding strategy.}
    \label{fig:se2_pipeline}
\end{figure*}

In this work, we propose an SE(2) Navigation Mesh (SE(2) NavMesh) representation for global navigation of ground robots in complex 3D environments. SE(2) NavMesh extends the classical NavMesh by incorporating the robot's yaw into navigation mesh construction, enabling more accurate traversability estimation and navigation in constrained scenes, as shown in~\Cref{fig:se2_navmesh}. The yaw-encoded structure allows global path planning to explicitly consider both the robot's position and heading. Since the robot's motion is constrained to the surface of the navigation mesh polygons, the resulting navigation space has two translational ($x, y$) and one rotational (yaw) degree of freedom, motivating the SE(2) naming. Built on this representation, we develop an A*-String Pulling-A* (ASA) pathfinding strategy that leverages the yaw-aware layered structure of the SE(2) NavMesh to hierarchically optimize robot position and yaw. We further introduce an online SE(2) NavMesh generation framework that incrementally constructs a navigation mesh from environment geometry reconstructed in real time from onboard sensor observations. We validate the proposed approaches through experiments in both simulated and real-world environments. The main contributions of this work are summarized as follows:
\begin{itemize}
    \item We propose SE(2) NavMesh, a novel representation that encodes yaw-dependent feasibility of traversable regions. Using footprint masks for traversability estimation, SE(2) NavMesh represents traversable narrow structures more accurately than the classical NavMesh, which we demonstrate across different scenes.
    \item We propose a three-stage ASA pathfinding strategy for the SE(2) NavMesh that exploits the layered structure of the SE(2) NavMesh to hierarchically optimize robot position and yaw. The proposed strategy reduces the geometric path length by \SI{6.2}{\percent} and lowers the final path cost to \SI{87}{\percent} of the initial path cost.
    \item We present an online SE(2) NavMesh generation method that incrementally constructs and updates the navigation mesh from point cloud observations. Combined with concurrent geometry reconstruction, the method runs efficiently on onboard CPUs, enabling real-time updates for real-world deployment.
\end{itemize}
An overview of the proposed SE(2) NavMesh pipeline is illustrated in~\Cref{fig:se2_pipeline} and detailed in~\Cref{sec:method}. A project page with an interactive path planning explorer and videos of real-world experiments is available at \url{https://se2-navmesh.github.io/}. Our implementation will be released as open source to support reproducibility and future research.

\section{Related Work}

Point clouds are a common representation of complex environments, which can be constructed through prior scanning~\cite{wand2008processing} or registered online during robot operation~\cite{tuna2023x, jelavic2022open3d} with widely used 3D sensors such as LiDAR or RGB-D cameras. To enable ground robot navigation on point clouds, Liu~\cite{liu2015robotic} applied tensor voting~\cite{medioni2000tensor} to infer local geometry and defined a Riemannian metric-based motion cost that favors planar regions while penalizing rough terrain. The method performs global planning via Dijkstra~\cite{dijkstra1959note} search on a graph constructed by connecting each point to its $k$-nearest neighbors ($k$-NN). A key limitation of this approach is that it neglects the robot's geometry and traversal capabilities, such as maximum climbing height and inclination limits, potentially yielding infeasible paths. Krüsi et~al.~\cite{krusi2017driving} performed global motion planning directly on point clouds using sampling-based planners, including RRT~\cite{lavalle2001rapidly} and RRT*~\cite{karaman2011sampling}. Local terrain traversability was assessed by estimating surface orientation via principal component analysis (PCA) of neighboring points~\cite{hoppe1992surface} and evaluating roughness over robot footprint-sized planar patches. While sampling-based planners avoid exhaustive traversability estimation over the full point cloud, statistics-based traversability evaluation may fail to accurately capture traversability in complex terrain. More broadly, the lack of structural organization in point clouds makes traversability estimation challenging, as geometric surfaces and connectivity must be inferred through additional processing that is error-prone and computationally expensive.

3D occupancy maps discretize the environment into voxels classified as free, occupied, or unknown, represented using uniform voxel grids~\cite{niessner2013real, ren2024rog, tang2026memory} or memory-efficient octree structures~\cite{hornung2013octomap, duberg2020ufomap}. Frey et~al.~\cite{frey2022locomotion} trained a sparse 3D convolutional neural network (CNN) over diverse simulated terrains to predict voxel-wise traversability risks from volumetric occupancy maps for global navigation. This learned traversability model is coupled to a specific locomotion policy and robot platform, making it difficult to parameterize for different robots. Furthermore, the predicted traversability risks must be combined with volumetric collision checking for obstacle avoidance~\cite{chen2023smug}. Wang et~al.~\cite{wang2023towards} extracted traversable ground regions from point clouds according to robot-specific constraints and performed grid-based path planning using a motion cost that combines ESDF-based collision avoidance with terrain roughness estimated from local plane fitting. Since traversability is evaluated using local slope estimates between neighboring points, stairs may be misclassified as non-traversable, restricting navigation for stair-capable robots. Li et~al.~\cite{li2025real} represented the environment using an octree-based map with normal distributions transform (NDT) Gaussian distributions stored in occupied voxels~\cite{ahtiainen2017normal}. Traversability is estimated from terrain metrics, including roughness, slope, and observation sparsity, followed by global path planning on a traversable voxel graph. Although the implicit Gaussian representation improves terrain characterization, the richer map representation increases the computational cost of online map updates.

Compared with 3D occupancy maps, 2D grid maps offer a simpler representation. Multiple layers can encode different obstacle types or constraints, yielding layered 2D grid maps~\cite{lu2014layered}. Kim et~al.~\cite{kim2019uav} used an unmanned aerial vehicle to acquire an initial 3D point cloud, which was discretized into a 2D grid-based navigation map for subsequent ground robot navigation, with cell traversability determined by terrain slope and height clearance. Planar grids, however, cannot represent terrain geometry, limiting their applicability in complex environments.

2.5D elevation maps~\cite{fankhauser2014robot} represent terrain using continuous height values and have been widely adopted for legged robot navigation over complex terrain. Wellhausen et~al.~\cite{wellhausen2021rough} proposed a reachability-based planner in which foothold feasibility is predicted from elevation maps using a CNN, with a customized LazyPRM*~\cite{hauser2015lazy} and adaptive sampling to achieve real-time performance. Yang et~al.~\cite{yang2021real} introduced a neural network that predicts energy cost, traversal duration, and failure probability from local elevation maps and motion commands, integrated into a GPU-accelerated framework combining graph search and gradient-based optimization for real-time navigation. Although effective on uneven terrain, the 2.5D nature of elevation maps fundamentally precludes their use in multi-floor environments. Yang et~al.~\cite{Yang2025tomography} addressed this by representing 3D environments as multiple 2.5D tomographic slices, computing traversability costs from ground geometry and ground-to-ceiling clearance, and performing A* search across slices via inter-slice transitions. The approach still assumes a circumscribed circular footprint, which may be overly conservative in narrow passages.

Another line of research uses triangle meshes, which encode surface geometry and connectivity and can be reconstructed from point clouds, ESDFs, or TSDFs~\cite{oleynikova2017voxblox, wiemann2018surface}. OVPC Mesh~\cite{ruetz2019ovpc} represents the surrounding environment as a watertight 3D mesh constructed from visible point clouds, classifying traversable space based on surface normals and height variation within each mesh polygon, and uses RRT-connect~\cite{kuffner2000rrt} for local path planning. Pütz et~al.~\cite{puetz2021trianglemesh} computed continuous shortest-path vector fields directly on triangle meshes via wavefront propagation, enabling extraction of global paths by following the field toward the goal. Compared with point clouds, triangle meshes provide a more continuous and interpretable surface representation. Nevertheless, these approaches rely on purely surface-based traversability estimation, struggle with traversable structures featuring vertical surfaces, and incur high planning times when dense meshes are required to represent complex environments.

\begin{table}[t]
\centering
\footnotesize
\begin{threeparttable}
\caption{Capability comparison of ground robot navigation methods}
\label{tab:nav_methods_comparison}
\begin{tabular}{@{}l
                >{\centering\arraybackslash}p{0.30cm}
                >{\centering\arraybackslash}p{0.30cm}
                >{\centering\arraybackslash}p{0.30cm}
                >{\centering\arraybackslash}p{0.30cm}
                >{\centering\arraybackslash}p{0.30cm}
                >{\centering\arraybackslash}p{0.30cm}
                >{\centering\arraybackslash}p{1.40cm}
                @{}}\toprule
&  \rotatebox{30}{\makecell{Multi-level Env.}} 
&  \rotatebox{30}{\makecell{Stair Handling}}
&  \rotatebox{30}{\makecell{Yaw-dep. Trav.}} 
&  \rotatebox{30}{\makecell{Config. Footprint}}
&  \rotatebox{30}{\makecell{Config. Trav. Cap.}}
&  \rotatebox{30}{\makecell{Online Gen.}} 
\\
\midrule
Liu~\cite{liu2015robotic} & \cmark & \cmark & \xmark & \xmark & \xmark & \xmark & \\
Krüsi et~al.~\cite{krusi2017driving} & \cmark & \cmark & \xmark & \cmark & \cmark & \cmark & \\
Frey et~al.~\cite{frey2022locomotion} & \cmark & \cmark & \xmark & \xmark & \xmark & \cmark & \\
Wang et~al.~\cite{wang2023towards} & \cmark & \xmark & \xmark & \cmark & \cmark & \xmark & \\
Kim et~al.~\cite{kim2019uav} & \xmark & \xmark & \xmark & \cmark & \cmark & \xmark & \\
Li et~al.~\cite{li2025real} & \cmark & \cmark & \xmark & \cmark & \cmark & \cmark & \\
Wellhausen et~al.~\cite{wellhausen2021rough} & \xmark & \cmark & \xmark & \xmark & \xmark & \cmark & \\
Yang et~al.~\cite{yang2021real} & \xmark & \cmark & \cmark & \xmark & \xmark & \cmark & \\
Yang et~al.~\cite{Yang2025tomography} & \cmark & \cmark & \xmark & \cmark & \cmark & \xmark & \\
OVPC Mesh~\cite{ruetz2019ovpc} & \xmark & \xmark & \xmark & \cmark & \cmark & \cmark & \\
Pütz et~al.~\cite{puetz2021trianglemesh} & \cmark & \xmark & \xmark & \cmark & \cmark & \xmark & \\
GaitMesh~\cite{brandao2020gaitmesh} & \cmark & \cmark & \xmark & \cmark & \cmark & \xmark & \\
\hline
SE(2) NavMesh & \cmark & \cmark & \cmark & \cmark & \cmark & \cmark & \\
                      
\bottomrule
\end{tabular}

\begin{tablenotes}[flushleft]
    \scriptsize
    \item Multi-level Env.: Supports navigation across multi-level environments.
    \item Stair Handling: Handles discrete elevation changes such as stairs.
    \item Yaw-dep. Trav.: Evaluates traversability of target terrain across different yaws.
    \item Config. Footprint: Supports configurable robot footprint.
    \item Config. Trav. Cap.: Supports configurable robot traversal capabilities (maximum step height and slope angle).
    \item Online Gen.: Generates maps (and search graphs) incrementally during operation.
\end{tablenotes}
\end{threeparttable}
\end{table}

To overcome the limitations of direct navigation on triangle meshes, NavMeshes abstract traversable surfaces into connected polygons. Recast~\cite{mononen2014recastnav}, a widely adopted NavMesh generation framework, combines surface geometry from triangle meshes with voxel-based clearance and connectivity analysis to estimate traversability, producing polygons that represent traversable regions in multi-level environments and form a navigation graph supporting efficient A*~\cite{hart1968formal} search. GaitMesh~\cite{brandao2020gaitmesh} applies NavMeshes to ground robot navigation in large-scale multi-floor environments by integrating curvature-based terrain analysis and gait controller selection. However, these NavMesh representations assume yaw-independent traversability and approximate the robot as a cylinder, preventing accurate representation of narrow passages with yaw-dependent traversability. Furthermore, GaitMesh relies on an offline construction pipeline and does not support online generation from incremental observations. In this work, we propose an SE(2) NavMesh that incorporates robot yaw during mesh construction, enabling better traversability capture in narrow passages and supporting planning of both heading and motion direction. An online generation pipeline constructs maps from onboard sensing, after which the resulting navigation mesh supports multiple planning queries. \Cref{tab:nav_methods_comparison} shows that, compared with existing approaches, SE(2) NavMesh uniquely supports all listed navigation capabilities.

\section{Problem Formulation}
\label{sec:problem_formulation}
\begin{figure*}[t]
    \centering
    \begin{subfigure}{0.13\linewidth}
        \centering
        \includegraphics[width=\linewidth]{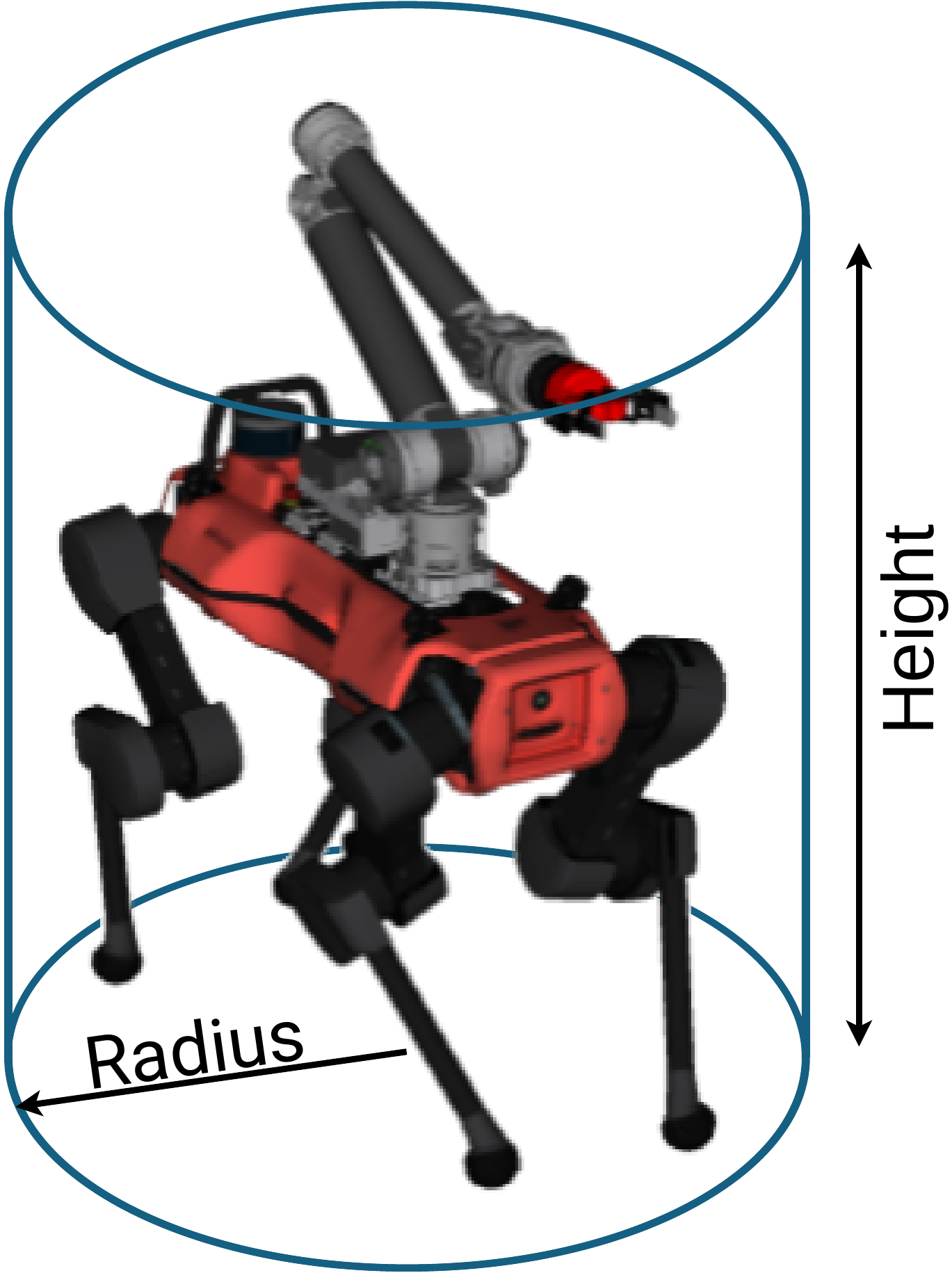}
        \caption{}
        \label{fig:cylindrical_approximation}      
    \end{subfigure}
    \begin{subfigure}{0.14\linewidth}
        \centering
        \includegraphics[width=\linewidth]{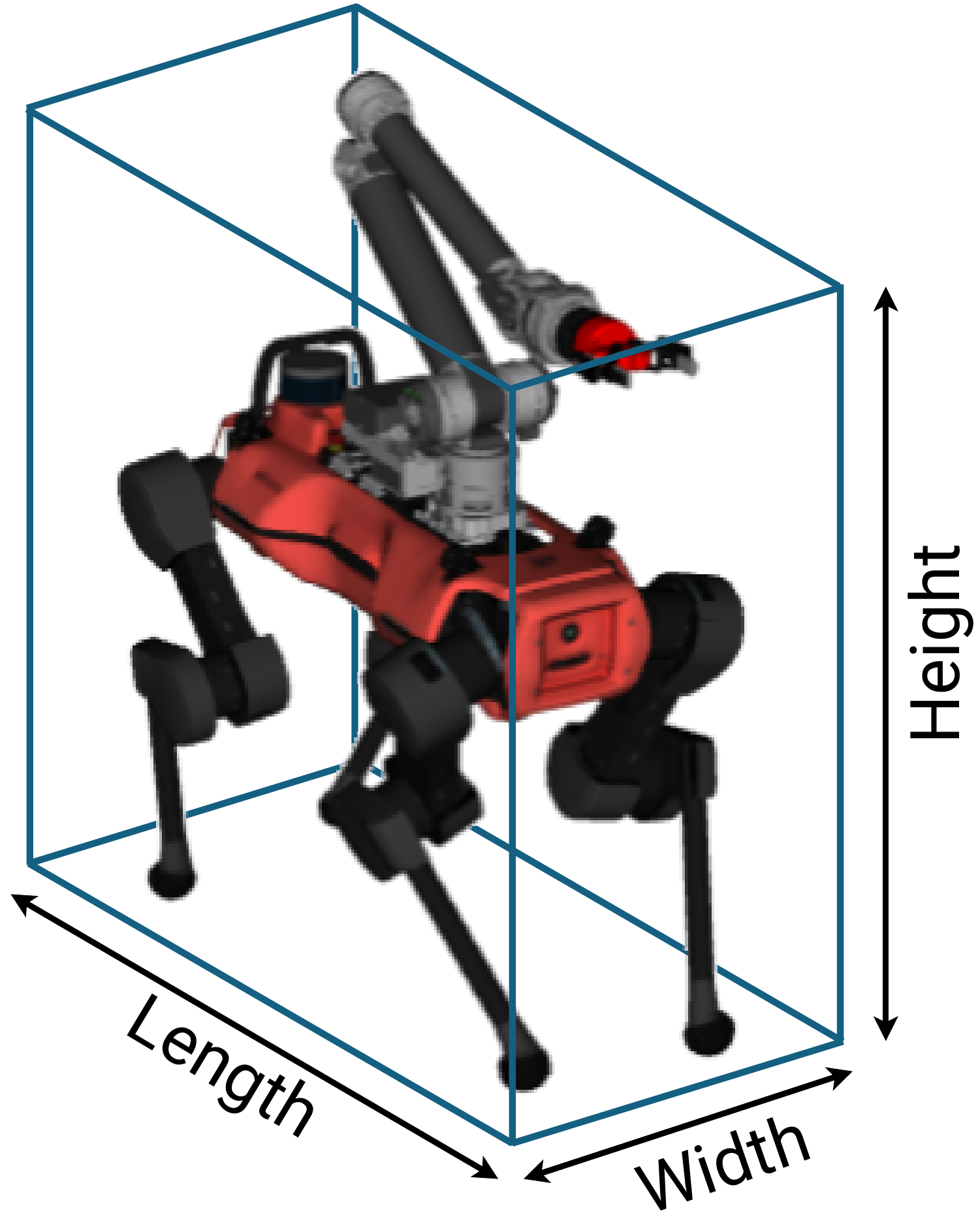}
        \caption{}
        \label{fig:cuboid_approximation}       
    \end{subfigure}
    \begin{subfigure}{0.7\linewidth}
        \centering
        \includegraphics[width=\linewidth]{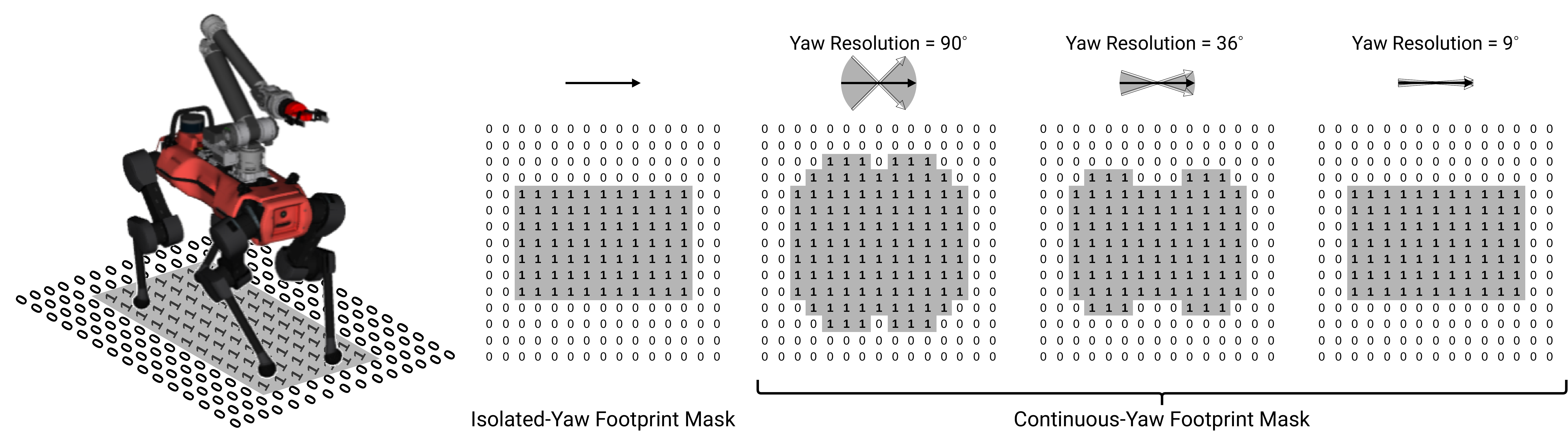}
        \caption{}
        \label{fig:footprint_mask}       
    \end{subfigure}

    \caption{Robot geometry approximations and footprint masks. (a) Cylindrical approximation. (b) Cuboid approximation. (c) Binary footprint masks, where 1 denotes cells occupied by the robot footprint and 0 denotes unoccupied cells.}
    \label{fig:footprint_mask_generation}
\end{figure*}

Consider a holonomic ground robot—typical of legged systems—capable of omnidirectional translation and yawing across traversable surfaces. Assume that the robot maintains a relatively fixed planar footprint and that its traversal capability parameters are given. The robot traverses a 3D environment represented as a polygonal surface mesh \(\mathset{M} \subset \mathbb{R}^3\). Its state is defined as
\begin{equation}
    \mathvec{x} = (\mathvec{p},\psi) \;,
\end{equation}
where \(\mathvec{p} \in \mathset{M}\) denotes the projection of the reference point of the robot base onto the nearest polygonal surface (along the direction of gravity) and \(\psi \in S^1\) denotes the yaw. Then, the configuration space of the robot is
\begin{equation}
    \mathset{X} = \mathset{M} \times S^1 \;.
\end{equation}
The robot's geometry and traversability characteristics further restrict the feasible configurations. The resulting traversable state space is defined as
\begin{equation}
    \mathset{X}_\mathrm{free} = \{(\mathvec{p},\psi) \in \mathset{X} \mid \mathvec{p} \in \mathset{P}_\mathrm{free}(\psi)\} \;,
\end{equation}
where \(\mathset{P}_\mathrm{free}(\psi) \subset \mathset{M}\) denotes the set of feasible positions for a fixed yaw. This formulation, therefore, captures the yaw-dependent traversable surfaces of the environment for the robot.

Our objective is to develop a representation that supports traversability estimation and pathfinding for this robot in a 3D environment.
In this context, traversability estimation is the task of deriving the \(\mathset{X}_\mathrm{free}\) from \(\mathset{M}\). Pathfinding is then reduced to the problem of computing a continuous path within \(\mathset{X}_\mathrm{free}\) between a given start state \(\mathvec{x}_\mathrm{start} \in \mathset{X}_\mathrm{free}\) and a goal state \(\mathvec{x}_\mathrm{goal} \in \mathset{X}_\mathrm{free}\). Crucially, this representation should also be capable of running online and being generated directly from sensor input. 

\section{Preliminaries}
\label{sec:preliminaries}

In this section, we review the foundational concepts of the classical NavMesh, following the formulation adopted in Recast~\cite{mononen2014recastnav}.

\subsection{Robot Model and Cylindrical Approximation}

In the NavMesh, the robot is modeled using four parameters that characterize its geometry and traversal capabilities. As illustrated in \Cref{fig:cylindrical_approximation}, a cylindrical approximation is employed to represent the robot's geometry, which is parameterized by the robot height \(h_\mathrm{robot}\) and the robot circumradius \(r_\mathrm{circ}\), defined as the radius of the minimum circumscribed circle of its footprint. The robot traversal capabilities are described by the maximum step height \(h_\mathrm{step}\) and the maximum slope angle \(\theta_\mathrm{climb}\). Given that the robot remains upright, the cylinder axis is always aligned with the direction of gravity. Under this modeling paradigm, the robot projects a circular footprint onto the ground.

\subsection{Traversable Regions}

\begin{figure}[t]
    \centering

    \begin{subfigure}{0.48\linewidth}
        \centering        \includegraphics[width=0.95\linewidth]{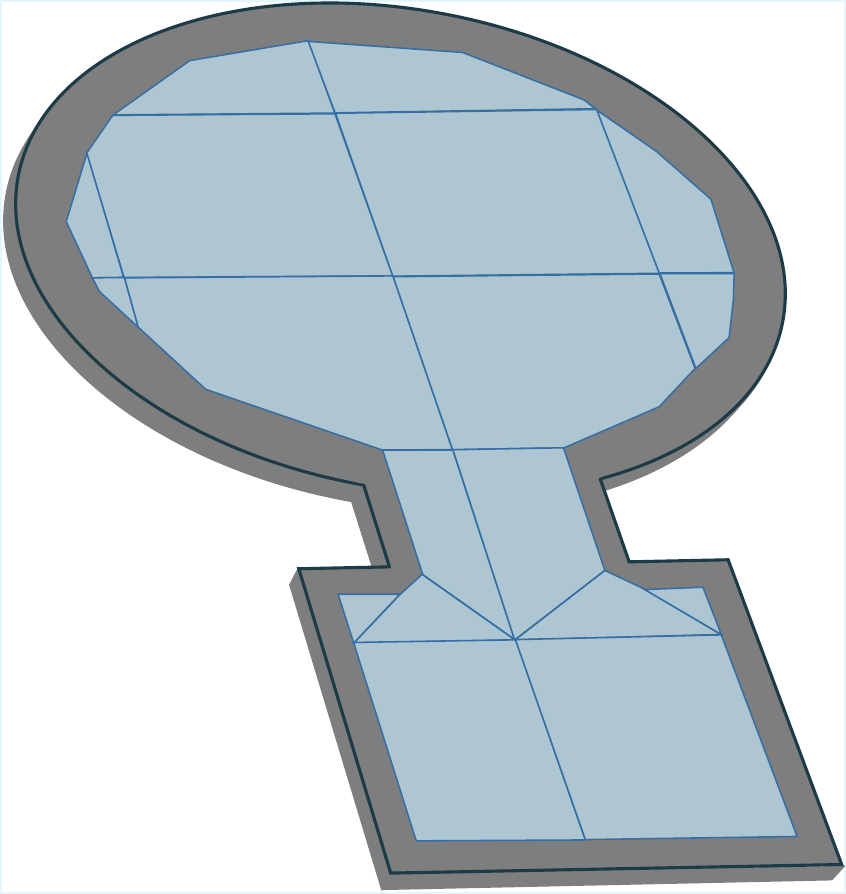}
        \caption{Traversable regions.}
        \label{fig:navmesh}
    \end{subfigure}
    \hfill
    \begin{subfigure}{0.48\linewidth}
        \centering        \includegraphics[width=0.85\linewidth]{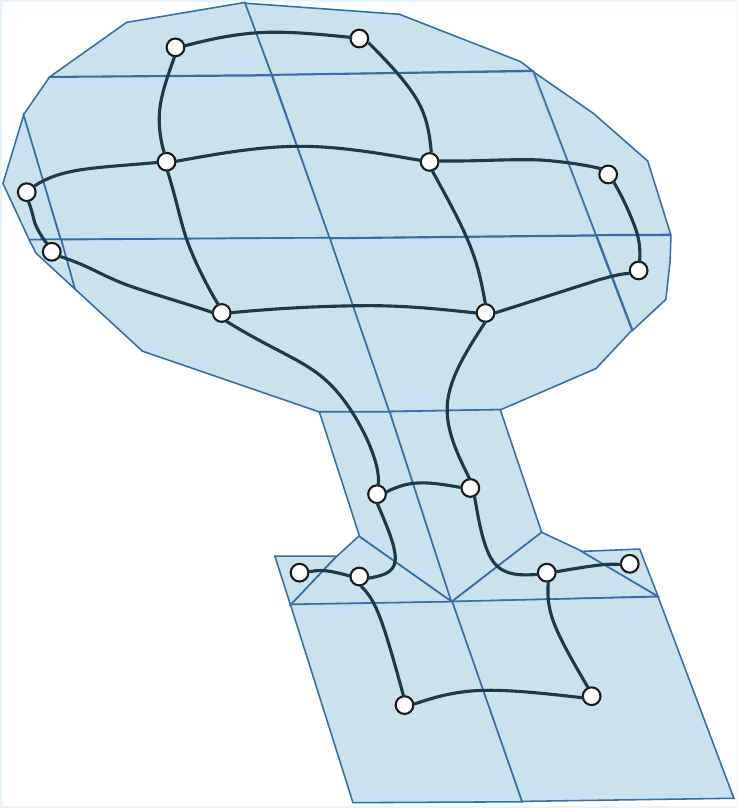}
        \caption{Graph.}
        \label{fig:navmesh_graph}
    \end{subfigure}
    \caption{Classical NavMesh representation and graph construction. (a) Traversable regions are represented as non-overlapping convex polygons. (b) The graph represents each polygon as a vertex, with edges encoding connectivity between adjacent polygons.}
    \label{fig:navmesh_to_graph}
\end{figure}

Since the NavMesh does not consider yaw during traversability estimation, it requires that a position be feasible for all possible yaws, corresponding to taking the intersection over all yaw-dependent feasible sets: 
\begin{equation}
    \bigcap_{\psi \in S^1} \mathset{P}_\mathrm{free}(\psi) \;.
\end{equation}
This provides a conservative representation of the traversable surfaces in the environment. 
The NavMesh partitions traversable positions and performs polygonal contour extraction, yielding a simplified representation as a set of traversable regions (see \Cref{fig:navmesh}):
\begin{equation}
    \mathset{R}_{\mathrm{trav}} = \{ \mathset{R}_1, \mathset{R}_2, \mathset{R}_3, \ldots \}.
\end{equation}
Each \(\mathset{R}_i \subset \mathset{M}\) denotes a set of traversable positions represented by a convex polygonal region.
Therefore, the set of traversable positions under the NavMesh representation can be written as
\begin{equation}
    \mathset{P}_{\mathrm{trav}} = \bigcup_{\mathset{R}_i \in \mathset{R}_{\mathrm{trav}}} \mathset{R}_i \; .
\end{equation}
We can further write the estimated traversable state space as
\begin{equation}
    \mathset{X}_{\mathrm{trav}} = \mathset{P}_{\mathrm{trav}} \times S^1.
\end{equation}
Due to the conservative estimation of traversable regions in the classical NavMesh, the induced state space satisfies
\begin{equation}
    \mathset{X}_{\mathrm{trav}} \subseteq \mathset{X}_{\mathrm{free}} .
\end{equation}
Consequently, feasible configurations that are valid only for specific yaws are not captured by this representation.

\subsection{Connectivity of Traversable Regions}

Connectivity is defined as a binary relation over traversable regions: two regions \(\mathset{R}_i\) and \(\mathset{R}_j\) are connected if their boundaries have overlapping segments. This connectivity relation can be written as:
\begin{equation}
    \Conn(\mathset{R}_i,\mathset{R}_j) =
    \begin{cases}
    1, & \text{if } \mathcal{H}^1\!\left(\partial \mathset{R}_i \cap \partial \mathset{R}_j \right) > 0 \\
    0, & \text{otherwise}
    \end{cases} ,
\end{equation}
where \(\partial \mathset{R}_i\) denotes the boundary of region \(\mathset{R}_i\) and $\mathcal{H}^1$ is the 1-dimensional Hausdorff measure. The connectivity definition implies that the robot can traverse freely between the two regions through the shared polygon edge segment.

\subsection{Graph}

Based on the defined connectivity relation, the set of traversable regions can be transformed into a graph representation that enables efficient pathfinding. Graph \(G = (\mathset{V},\mathset{E})\) describes how the robot can navigate between traversable regions. Each vertex \(v_i \in \mathset{V}\) corresponds to a traversable region \(\mathset{R}_i \in \mathset{R}_{\mathrm{trav}}\). An edge between two vertices exists if the corresponding traversable regions are connected, i.e., 
\begin{equation}
    (v_i,v_j) \in \mathset{E} \iff \Conn(\mathset{R}_i,\mathset{R}_j) = 1
\end{equation}
\Cref{fig:navmesh_graph} shows an example of converting the traversable regions and their connectivity into a graph. Pathfinding is achieved via an A* search conducted on the graph. Details regarding A* pathfinding on NavMesh can be found in~\cite{cui2011based}.

\section{Method}
\label{sec:method}

\Cref{fig:se2_pipeline} provides an overview of the proposed SE(2) NavMesh pipeline. This section first defines the components of SE(2) NavMesh and formalizes their relations, clarifying how SE(2) NavMesh extends the classical NavMesh. It then describes the offline generation method, the associated pathfinding procedure, and the online generation method.

\subsection{SE(2) NavMesh Definitions}

\begin{figure*}[t]
    \centering
    \includegraphics[width=0.98\linewidth]{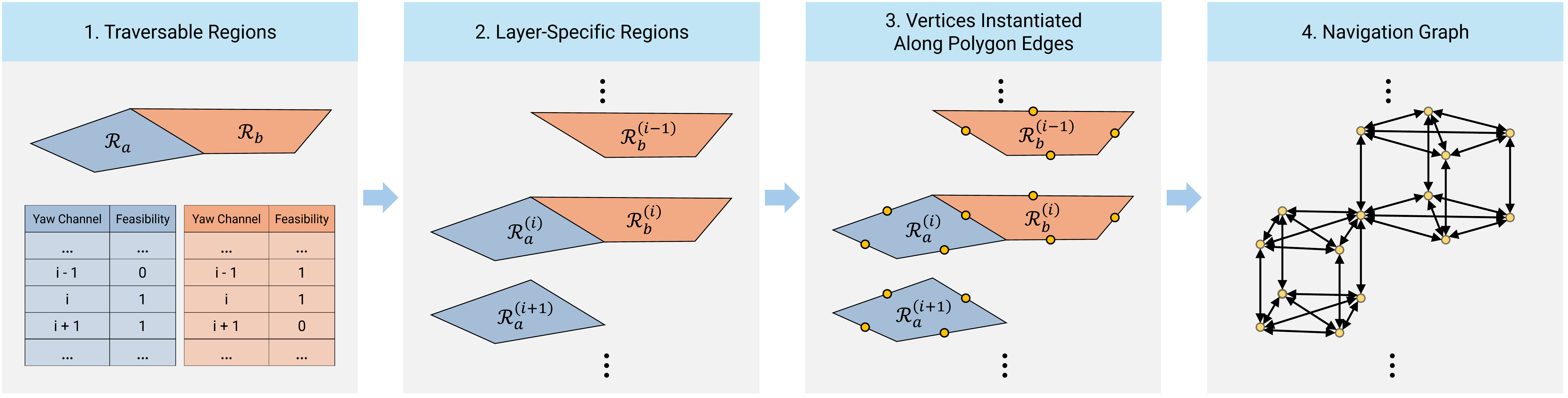}

    \caption{Generation of layer-specific regions and the navigation graph. Each yaw channel corresponds to a yaw-specific traversability layer. According to the feasible yaw channel sets of traversable regions, layer-specific regions are generated. Graph vertices are instantiated along the polygon edges of the layer-specific regions, and connectivity is established within and across layers to construct the navigation graph.}
    \label{fig:se2_layer}
\end{figure*}

\subsubsection{Robot Model and Cuboid Approximation}

In this work, we demonstrate the framework using a legged robot with a rectangular footprint. Importantly, the framework itself is not constrained by this choice and applies to ground robots with arbitrary shapes.
We describe the geometry of the robot by its length \(l_\mathrm{robot}\), width \(w_\mathrm{robot}\), and height \(h_\mathrm{robot}\), as illustrated in~\Cref{fig:cuboid_approximation}. Compared with the cylindrical approximation, the cuboid approximation provides a more compact representation of the robot. In the same way that Recast adopts a cylindrical approximation, the height direction of the cuboid is aligned with the direction of gravity. The traversal capabilities of the robot are characterized by the maximum step height \(h_\mathrm{step}\) and the maximum slope angle \(\theta_\mathrm{climb}\).

\subsubsection{Yaw Channel}

The continuous yaw interval \([0,2\pi)\) is uniformly discretized into \(N_{\Psi}\) yaw angles, forming the set 
\begin{equation}
    \Psi = \{\psi_1,\psi_2,\psi_3,\ldots,\psi_{N_{\Psi}}\},
\end{equation}
where each \(\psi_i\) corresponds to a robot heading
\begin{equation}
     \psi_i = i \cdot \frac{2\pi}{N_{\Psi}} \;.
\end{equation}
Each heading \(\psi_i\) is associated with a yaw channel indexed by \(i\). Correspondingly, the set of yaw channels is defined as
\begin{equation}
    \mathset{I}_\Psi = \{1,2,\ldots,N_{\Psi}\}.
\end{equation}
For a potential position \(\mathvec{p}\in \mathset{M}\), the feasibility of the \(i\)-th yaw channel is:
\begin{equation}
    C_i(\mathvec{p}) =
    \begin{cases}
        1, & \text{if robot can occupy } \mathvec{p} \text{ with yaw } \psi_i \\
        0, & \text{otherwise}
    \end{cases} .
\end{equation}

\subsubsection{Footprint Mask}

As illustrated in~\Cref{fig:footprint_mask}, the footprint mask is a binary matrix representing the spatial occupancy of the robot footprint on the horizontal plane (e.g., a rectangular footprint for our cuboid-shaped legged robots). Each cell of the mask has a value of 1 if the corresponding position is occupied by the footprint, and 0 otherwise. One cell within the occupied region is designated as the reference cell of the robot base, typically chosen as the center cell.

Depending on the intended use, a footprint mask can be generated either for an isolated yaw angle or for a continuous range of yaw angles. In the former case, the footprint mask, termed the isolated-yaw footprint mask, represents the planar occupancy of the robot associated with a single yaw angle. Such a mask is suitable for collision checking at a fixed robot yaw angle during path planning. In the latter case, the footprint mask, termed the continuous-yaw footprint mask, captures the union of the occupied areas swept by the robot footprint over the yaw interval
\begin{equation}
    \left[\psi_i-\frac{\psi_\mathrm{interval}}{2}, \;\psi_i+\frac{\psi_\mathrm{interval}}{2}\right),\; \psi_\mathrm{interval} = \frac{2\pi}{N_{\Psi}} \;.
    \label{eq:angle_interval}
\end{equation}
If a position is feasible under the continuous-yaw footprint masks of two adjacent yaw angles, then a safe rotation between these two headings is guaranteed.

\subsubsection{Safe and Restricted Traversable Regions}
\label{subsubsec:safe_restricted_regions}

For each region \(\mathset{R}_j \in \mathset{R}_{\mathrm{trav}}\), yaw feasibility is defined to be uniform within the region, such that all positions in \(\mathset{R}_j\) share the same feasible yaw channels. We therefore define the feasible yaw channel set \(\mathset{I}(\mathset{R}_j) \subseteq \mathset{I}_\Psi\) as the set of channels that permit traversal within \(\mathset{R}_j\). 
Formally, for any position \(\mathvec{p} \in \mathset{R}_j\) and channel \(i \in \mathset{I}_\Psi\),
\begin{equation}
    C_i(\mathvec{p}) =
    \begin{cases}
        1, & i \in \mathset{I}(\mathset{R}_j) \\
        0, & i \notin \mathset{I}(\mathset{R}_j)
    \end{cases}.
\end{equation}

Depending on the relationship between \(\mathset{I}(\mathset{R}_j)\) and \(\mathset{I}_\Psi\), we partition the overall traversable space into two distinct subsets: safe (\(\mathset{R}_{\mathrm{safe}}\)) and restricted (\(\mathset{R}_{\mathrm{restricted}}\)) traversable regions, such that
\begin{equation}
    \mathset{R}_{\mathrm{trav}} = \mathset{R}_{\mathrm{safe}} \cup \mathset{R}_{\mathrm{restricted}}\, .
\end{equation}
A region \(\mathset{R}_j\) is classified as safe if the robot can traverse it under all discretized yaw angles,
\begin{equation}
    \mathset{I}(\mathset{R}_j) = \mathset{I}_\Psi,
\end{equation}
and is classified as restricted if traversal is feasible only under a subset of yaw angles,
\begin{equation}
    \emptyset \subsetneq \mathset{I}(\mathset{R}_j) \subsetneq \mathset{I}_\Psi.
\end{equation}

\subsubsection{Yaw-Specific Traversability Layer}

For each yaw channel \(i \in \mathset{I}_\Psi\), we define a yaw-specific traversability layer \(\mathset{L}_i\). Each layer represents the subset of traversable regions where the robot can traverse under this specific heading: 
\begin{equation}
    \mathset{L}_i = \{\mathset{R}_j \in \mathset{R}_\mathrm{trav} \mid i \in \mathset{I}(\mathset{R}_j) \}.
\end{equation}
The relationship between the yaw-specific layers and the safe/restricted traversable regions can be expressed as 
\begin{subequations}
\begin{align}
    \mathset{R}_\mathrm{trav} &= \bigcup_{i \in \mathset{I}_\Psi} \mathset{L}_i, \\
    \mathset{R}_\mathrm{safe} &= \bigcap_{i \in \mathset{I}_\Psi} \mathset{L}_i, \\
    \mathset{R}_\mathrm{restricted} &= \bigcup_{i \in \mathset{I}_\Psi} \mathset{L}_i \setminus \bigcap_{i \in \mathset{I}_\Psi} \mathset{L}_i .
\end{align}
\end{subequations}
To distinguish the occurrences of a traversable region \(\mathset{R}_j \in \mathset{R}_{\mathrm{trav}}\) across different yaw-specific traversability layers, we denote by \(\mathset{R}^{(i)}_j\) its layer-specific region in layer \(\mathset{L}_i\), as illustrated in~\Cref{fig:se2_layer}. The element \(\mathset{R}^{(i)}_j\) exists if and only if \(\mathset{R}_j\) is feasible at yaw angle \(\psi_i\). 

\subsubsection{Translational Connectivity and Rotational Connectivity}
\label{subsubsec:transaltional_rotational_connectivity}

When a robot moves between a restricted traversable region and another traversable region, it must carefully choose its yaw, as the yaw feasibility of the two regions may not coincide, and it might not be able to maintain a fixed heading. Consequently, connectivity in SE(2) NavMesh extends the standard NavMesh definition by incorporating the yaw angle, categorizing it into two types: translational and rotational. 
Translational connectivity is defined between two layer-specific regions that lie in the same yaw-specific traversability layer, corresponding to the robot's motion while keeping the fixed yaw.
Specifically, two layer-specific regions $\mathset{R}_a^{(i)}$ and $\mathset{R}_b^{(i)}$ from layer $\mathset{L}_i$ are translationally connected if \(\mathset{R}_a\) and \(\mathset{R}_b\) have overlapping polygon edges:
\begin{equation}
    \Conn_\mathrm{trans}(\mathset{R}_a^{(i)}, \mathset{R}_b^{(i)}) =
    \begin{cases}
    1, & \text{if } \mathcal{H}^1\!\left(\partial \mathset{R}_a \cap \partial \mathset{R}_b \right) > 0 \\
    0, & \text{otherwise}
    \end{cases} .
    \label{eq:trans_conn}
\end{equation}
Rotational connectivity is defined between different layer-specific regions of the same traversable region. It represents the robot's ability to rotate in-place within a region \(\mathset R\) while remaining collision-free.
Specifically, for a traversable region \(\mathset{R}_k\), two of its layer-specific regions \(\mathset{R}_k^{(i)}\) and \(\mathset{R}_k^{(j)}\) are considered rotationally connected if their corresponding yaw angles are adjacent:
\begin{equation}
    \Conn_\mathrm{rot}(\mathset{R}_k^{(i)},\mathset{R}_k^{(j)}) =
    \begin{cases}
    1, & \text{if } j \equiv i \pm 1 \pmod{N_{\Psi}} \\
    0, & \text{otherwise}
    \end{cases}
    \label{eq:rot_conn}
\end{equation}
Due to the periodicity of yaw, adjacency is interpreted cyclically, hence the use of the modulo operation.

\subsubsection{Navigation Graph}
\label{subsubsec:se2_graph}

For pathfinding purposes, we construct a navigation graph \(G=(\mathset{V},\mathset{E})\) based on layer-specific regions of traversable regions, as illustrated in~\Cref{fig:se2_layer}. 
For each polygon edge of every traversable region, a representative geometric point, such as its midpoint, is chosen and used to instantiate a vertex in each corresponding layer-specific region.
Vertices within the same layer share the same yaw attribute and are connected by translational edges \(\mathset{E}_\mathrm{trans}\). Vertices at the same spatial location in adjacent yaw-specific layers are connected by rotational edges \(\mathset{E}_\mathrm{rot}\), representing changes in the robot's heading. 
We define the edge cost as the time required for the robot to execute the corresponding motion.

\subsection{Offline Generation of SE(2) NavMesh}
\label{subsec:offline_generation}

The offline pipeline generates an SE(2) NavMesh from the input triangle mesh. Two sets of parameters are needed for the generation process:
\begin{itemize}
    \item \textit{Robot Parameters}: length ($l_\mathrm{robot}$), width ($w_\mathrm{robot}$), height ($h_\mathrm{robot}$), maximum step height ($h_\mathrm{step}$), and maximum slope angle ($\theta_\mathrm{climb}$).
    \item \textit{Map Configuration Settings}: the map voxelization resolution for length, width, and height ($s_\mathrm{voxel} \times s_\mathrm{voxel} \times h_\mathrm{voxel}$), and the number of yaw layers ($N_{\Psi}$).
\end{itemize}
The generation process first constructs footprint masks for discretized yaw angles. The environment is partitioned into tiles and converted into voxels that represent the local terrain. Yaw feasibility is then evaluated on these terrain voxels to determine traversable states under different headings. Finally, traversable voxels are aggregated to generate convex polygonal regions, which are merged across tiles to form the global SE(2) NavMesh together with its navigation graph.

\subsubsection{Footprint Masks Generation}

We use continuous-yaw footprint masks to ensure safe rotation. Each footprint mask, as shown in~\Cref{fig:footprint_mask}, is defined on a grid with a cell size of \(s_\mathrm{voxel} \times s_\mathrm{voxel}\), matching the resolution used for map voxelization.

\subsubsection{Tile Partitioning}
\label{subsubsec:tile_paritioning}

\begin{figure}[t]
    \centering
    \includegraphics[width=0.7\linewidth]{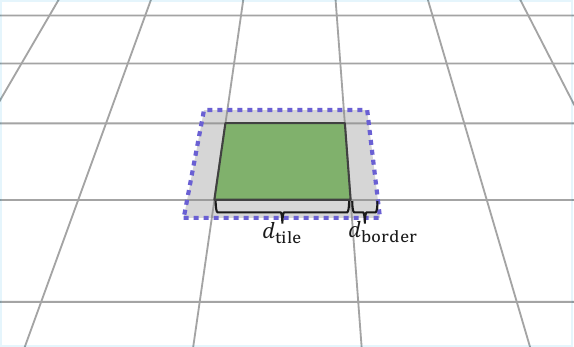}
    \caption{Tile-based SE(2) NavMesh generation. Generating the SE(2) NavMesh for a target tile (green) requires geometry from the tile itself and its neighboring area, shown as the gray region enclosed by the purple dashed line.}
    \label{fig:tile_mesh_build}
\end{figure}

The TileMesh approach from Recast~\cite{mononen2014recastnav} is adopted to generate SE(2) NavMesh. As illustrated in~\Cref{fig:tile_mesh_build}, the map is partitioned into square tiles on the horizontal plane, with each tile spanning \(d_\mathrm{tile} \times d_\mathrm{tile}\) voxels along the horizontal dimensions and being responsible for generating the SE(2) NavMesh for all height levels within its horizontal extent. This partitioning enables parallel processing, as each tile can be generated independently using only a local subset of the map geometry as input. This subset covers a square area of \((d_\mathrm{tile} + 2d_\mathrm{border})^2\) voxels, where \(d_\mathrm{border}\)
equals the circumradius of the robot footprint in voxels plus an additional safety margin of several voxels. This margin is required to correctly evaluate traversability near the tile border. All subsequent steps of the offline SE(2) NavMesh generation pipeline are carried out on a per-tile basis.

\subsubsection{Map Voxelization and Walkable Voxel Annotation}
\label{subsubsec:method_offline_map_voxelization}

The input triangle mesh is voxelized. Vertically connected voxels at the same planar position are merged, and we refer to them as \emph{solid spans}. The empty intervals between solid spans form the \emph{free space spans}. Based on surface geometry, robot height, and step constraints, each free space span is classified as walkable or non-walkable. The bottommost voxels of free space spans represent the terrain surfaces and collectively form a \emph{tile terrain voxel map}, which serves as the basis for the subsequent steps.

\subsubsection{Yaw Feasibility Evaluation}
\label{subsubsec:method_offline_yaw_feasibility}

On the tile terrain voxel map, a distance map is computed to represent the distance from each voxel to the nearest locally invalid voxel. A voxel is treated as locally invalid if it is non-walkable, adjacent to non-walkable voxels or unobserved regions, or if the height difference to any of its four neighboring voxels exceeds the traversability constraints. Locally invalid voxels are assigned a distance value of 0. Based on the distance map, voxels are further classified into three categories analogously to the traversable regions: safe, restricted, and inaccessible. A safe voxel can accommodate the robot footprint for any yaw, while a restricted voxel only supports the robot under a subset of yaw angles. An inaccessible voxel is too close to locally invalid voxels for the robot to reach.

The complete voxel classification procedure is summarized in~\Cref{alg:voxel_classification}. Under the rectangular footprint approximation, the footprint inradius \(r_\mathrm{in}\) and circumradius \(r_\mathrm{circ}\) are given by:
\begin{equation}
    r_\mathrm{in} = \frac{\min(l_\mathrm{robot},w_\mathrm{robot})}{2},
\end{equation}

\begin{equation}
    r_\mathrm{circ} = \frac{\sqrt{l_\mathrm{robot}^2 + w_\mathrm{robot}^2}}{2} \; .
\end{equation}

\begin{algorithm}[h]
    \caption{Voxel Classification}
    \label{alg:voxel_classification}
    \begin{algorithmic}[1]
        \renewcommand{\algorithmicrequire}{\textbf{Input:}}
        \renewcommand{\algorithmicensure}{\textbf{Output:}}
    
        \Require Tile terrain voxel map $\mathset{W}_\mathrm{tile}$, distance map $D$, yaw channel indices set $\mathset{I}_\Psi$, footprint masks $\{M_i\}_{i \in \mathset{I}_\Psi}$, footprint inradius $r_\mathrm{in}$, footprint circumradius $r_\mathrm{circ}$
        \Ensure Feasible yaw channel set $\mathset{I}(v)$ and voxel classification $\Class(v)$ for all $v \in \mathset{W}_\mathrm{tile}$

        \For{\textbf{each} voxel $v \in \mathset{W}_\mathrm{tile}$}
            \State $\mathset{I}(v) \gets \emptyset$ \Comment{Initialize feasible yaw set}
            \State $d \gets D(v) $
            \If{$d < r_\mathrm{in}$}
                \State $\Class(v) \gets \text{\textsc{Inaccessible}}$
            \ElsIf{$d \ge r_\mathrm{circ}$}
                \State $\Class(v) \gets \text{\textsc{Safe}}$
                \State $\mathset{I}(v) \gets \mathset{I}_\Psi$
            \Else
                \For{\textbf{each} yaw index $i \in \mathset{I}_\Psi$}
                    \State \textsc{AlignReferenceCell}$(M_i, v)$
                    \If{\textsc{MaskFeasible}$(M_i, v, \mathset{W}_\mathrm{tile})$}
                        \State $\mathset{I}(v) \gets \mathset{I}(v) \cup \{i\}$
                    \EndIf
                \EndFor

                \If{$\mathset{I}(v) = \emptyset$}
                    \State $\Class(v) \gets \text{\textsc{Inaccessible}}$
                \Else
                    \State $\Class(v) \gets \text{\textsc{Restricted}}$
                \EndIf  
            \EndIf
        \EndFor
    \end{algorithmic}
\end{algorithm}
Based on the distance map, voxels whose distance to locally invalid voxels is smaller than the footprint inradius are directly classified as inaccessible, while voxels whose distance exceeds or equals the footprint circumradius are classified as safe. However, distance information alone is insufficient for voxels in the intermediate range between the inradius and circumradius. These voxels therefore require additional yaw feasibility evaluation using the precomputed continuous-yaw footprint masks. For a given yaw channel, feasibility at the voxel level is determined by whether all occupied cells in the footprint mask correspond to walkable voxels. To perform this check, the footprint mask is first aligned with the candidate voxel by translating the mask so that the designated reference cell coincides with the voxel. Starting from this reference cell, a breadth-first search is performed over all occupied cells in the mask. The feasibility check succeeds if and only if all visited cells correspond to walkable voxels in the terrain voxel map.

\subsubsection{Region Polygon Generation}
\label{subsubsec:method_offline_polygon_generation}

Based on the distance field, a watershed algorithm~\cite{beucher1979use} is applied to partition safe voxels and restricted voxels into non-overlapping regions. During the partitioning process, only voxels with identical feasible yaw channel sets are allowed to be grouped into the same region. The resulting regions are generally irregular in shape and are triangulated via the ear clipping algorithm~\cite{eberly2008triangulation} and subsequently merged into convex polygonal regions following the standard pipeline in~\cite{mononen2014recastnav}. 
This ensures straight-line reachability within each region, facilitating efficient path planning and path smoothing. These convex polygonal regions are categorized into safe and restricted traversable regions according to their feasible yaw channel sets, as defined in~\Cref{subsubsec:safe_restricted_regions}. The regions constructed within each tile are merged into a global structure, where translational and rotational connectivity are established according to~\cref{subsubsec:transaltional_rotational_connectivity}.

\subsection{Pathfinding}
\label{subsec:method_pathfinding}

\begin{figure*}[t]
    \centering
    \includegraphics[width=0.95\textwidth]{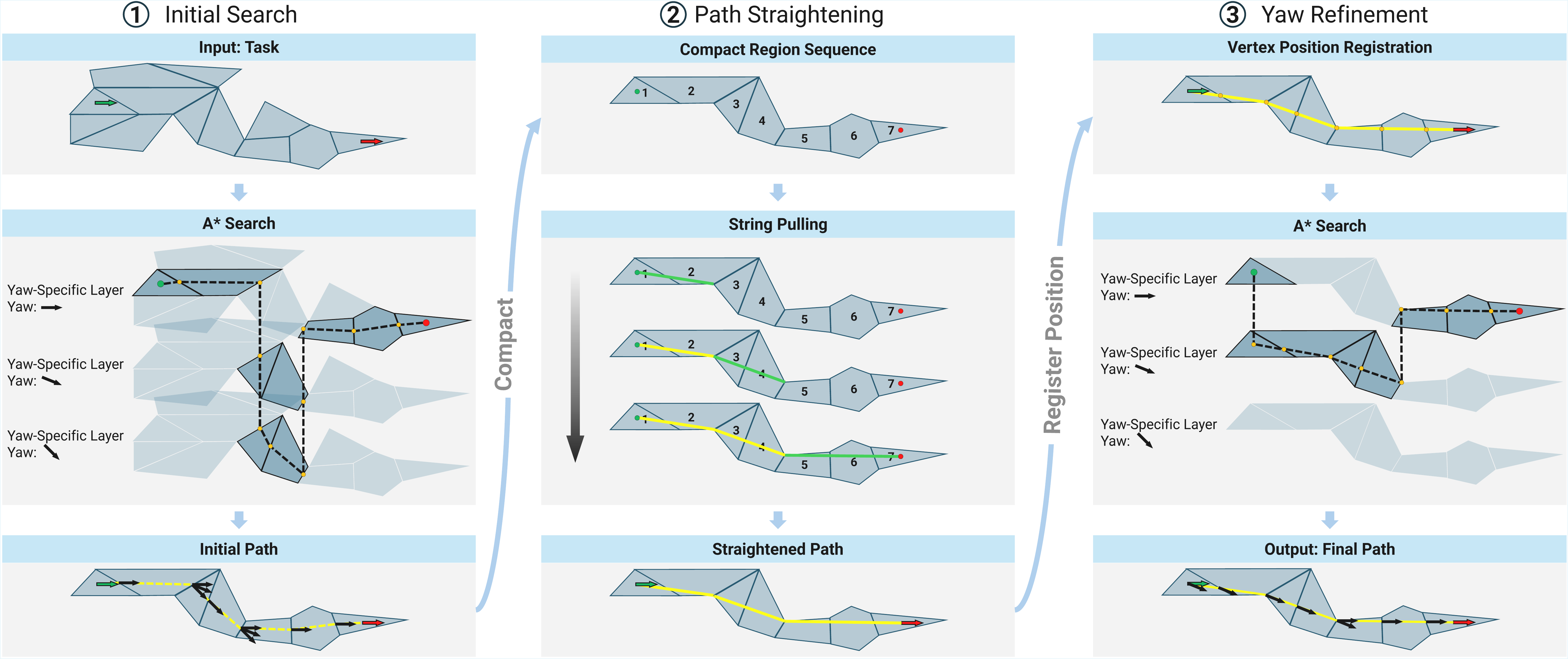}
    \caption{Overview of the ASA pathfinding strategy. The method consists of three stages: (1) an initial A* search that obtains a feasible path and layer-specific region sequence, (2) string pulling that straightens the path, and (3) yaw refinement through a second A* search that optimizes robot headings while respecting the straightened path constraints.}
    \label{fig:asa_pathfinding}
\end{figure*}

We propose a three-stage pathfinding strategy for the SE(2) NavMesh, termed ASA (A*-String Pulling-A*), as illustrated in~\Cref{fig:asa_pathfinding}. The strategy is structured as follows: 1) Initial Search: An A* algorithm computes an initial path and its corresponding layer-specific region sequence. 2) Path Straightening: A string-pulling method geometrically refines the path to produce straightened waypoints. 3) Yaw Refinement: A second A* search optimizes the robot's heading over the refined positions, constrained by the compact region sequence. The specific formulations for each stage are detailed below.

\subsubsection{Initial Search}
\label{subsubsec:initial_Search}

Given a start state \(\mathvec{x}_\mathrm{start} \in \mathset{X}_\mathrm{free}\) and a goal state \( \mathvec{x}_\mathrm{goal} \in \mathset{X}_\mathrm{free}\), we first identify the traversable regions \(\mathset{R}_\mathrm{start}\) and \(\mathset{R}_\mathrm{goal}\) that contain the spatial positions \(\mathvec{p}_\mathrm{start}\) and \(\mathvec{p}_\mathrm{goal}\). The yaw \(\psi_\mathrm{start}\) and \(\psi_\mathrm{goal}\) are then used to determine the corresponding yaw-specific layer indices \(l_\mathrm{start}\) and \(l_\mathrm{goal}\), locating the layer-specific regions \(\mathset{R}_\mathrm{start}^{(l_\mathrm{start})}\) and \(\mathset{R}_\mathrm{goal}^{(l_\mathrm{goal})}\). The navigation graph is constructed using vertices predefined at the midpoints of polygon edges for each layer-specific region. At the start and goal positions, temporary query vertices are instantiated at the same spatial location for all corresponding layer-specific regions. Vertices are connected according to the graph construction defined in~\Cref{subsubsec:se2_graph}.

During the A* search, the cost function is defined according to the type of edge connecting two states $\mathvec{x}_a = (\mathvec{p}_a, \psi_a)$ and $\mathvec{x}_b = (\mathvec{p}_b, \psi_b)$. We define the cost as the estimated traversal time between two states: The translational cost is determined by the robot's longitudinal and lateral velocities, while the rotational cost accounts for the robot's yaw rate.
A translational edge at the $i$-th layer corresponds to a movement from $(\mathvec{p}_a, \psi_i)$ to $(\mathvec{p}_b, \psi_i)$ while maintaining the current yaw $\psi_i$. Let \(\Delta \mathvec{p}_{xy}\) denote the horizontal projection of the displacement between the two states, obtained from the \((x,y)\) components of \(\mathvec{p}_b-\mathvec{p}_a\). The robot's local longitudinal and lateral directions are represented by the unit vectors:
\begin{equation}
    \mathvec{n}_\mathrm{long} = \begin{bmatrix} \cos \psi_i \\ \sin \psi_i \end{bmatrix}, \quad \mathvec{n}_\mathrm{lat} = \begin{bmatrix} -\sin \psi_i \\ \cos \psi_i \end{bmatrix} .
\end{equation}
By projecting the displacement vector onto these local axes, the translational cost is defined as:
\begin{equation}
    \Cost(\mathvec{x}_a, \mathvec{x}_b) = \frac{|\Delta \mathvec{p}_{xy}  \cdot \mathvec{n}_\mathrm{long}|}{v_\mathrm{long}} + \frac{|\Delta \mathvec{p}_{xy}  \cdot \mathvec{n}_\mathrm{lat}|}{v_\mathrm{lat}} ,
\end{equation}
where $v_\mathrm{long}$ and $v_\mathrm{lat}$ denote the robot's longitudinal and lateral velocities, respectively.
If a rotational edge exists between $\mathvec{x}_a$ and $\mathvec{x}_b$, the motion corresponds to an in-place rotation ($\mathvec{p}_a = \mathvec{p}_b$) of angle $\psi_\mathrm{interval}$ (defined in~\Cref{eq:angle_interval}) between two adjacent layers. The corresponding cost is defined as:
\begin{equation}
    \Cost(\mathvec{x}_a, \mathvec{x}_b) = \frac{\psi_\mathrm{interval}}{\omega} = \frac{2\pi}{N_\Psi \omega} ,
\end{equation}
where $\omega$ denotes the robot's yaw rate.

The heuristic function used in the A* search is defined as:
\begin{multline}
    \Heur(\mathvec{x}_j) = \frac{\lVert \mathvec{p}_\mathrm{goal} - \mathvec{p}_j \rVert_2}{\max \{v_\mathrm{long}, v_\mathrm{lat}\}} \\
    + \frac{\min \{|\psi_\mathrm{goal} - \psi_j|, 2\pi - |\psi_\mathrm{goal} - \psi_j|\}}{\omega} ,
\end{multline}
which estimates the remaining time to reach the goal by combining a lower bound on the translational travel time and a lower bound on the rotational adjustment time.

The A* search returns a sequence of states 
\begin{equation}
    \tau_\mathrm{init} = \left( \mathvec{x}_\mathrm{start}, \mathvec{x}_1, \mathvec{x}_2, \ldots, \mathvec{x}_\mathrm{goal} \right), \quad \mathvec{x}_k = (\mathvec{p}_k, \psi_k^\tau) \;, 
\end{equation}
where each state explicitly encodes both the spatial position and the yaw of the robot. This state sequence spans multiple yaw-specific layers and provides a feasible connection between the start and goal states.

Each state \(\mathvec{x}_i\) is associated with a layer-specific region \(\mathset{R}_{k_i}^{(l_i)}\). Mapping the states in \(\tau\) to their corresponding layer-specific regions yields an ordered sequence
\begin{equation}
    \Pi_\mathrm{init} = \left( \mathset{R}_\mathrm{start}^{(l_\mathrm{start})}, \mathset{R}_{k_1}^{(l_1)}, \mathset{R}_{k_2}^{(l_2)}, \ldots, \mathset{R}_\mathrm{goal}^{(l_\mathrm{goal})} \right) ,
\end{equation}
which describes the layer-specific regions traversed.

\subsubsection{Path Straightening}

The sequence of states \(\tau_\mathrm{init}\) is generally not length-optimal in the Euclidean sense, since it is constrained to pass through the midpoints of the polygon edges. To reduce the resulting zigzag behavior, we perform geometric path refinement using string pulling. The sequence \(\Pi_\mathrm{init}\) is first projected onto the original traversable regions by discarding yaw-specific layer information. Consecutive occurrences of the same regions are then merged to obtain a compact region sequence. For example, the sequence
\begin{equation}
    \left( \mathset{R}_a^{(2)}, \mathset{R}_b^{(2)}, \mathset{R}_b^{(3)}, \mathset{R}_b^{(4)}, \mathset{R}_c^{(4)} \right)
\end{equation}
is compacted to
\begin{equation}
    \Pi_\mathrm{compact} = \left( \mathset{R}_a,\mathset{R}_b,\mathset{R}_c \right) \;.
\end{equation}
The compact region sequence \(\Pi_\mathrm{compact}\) defines a continuous polygonal corridor. Within this corridor, string pulling is applied to compute a shortest path in the Euclidean sense, connecting the start and goal. We denote the resulting straightened path as
\begin{equation}
    \gamma =  \left( \mathvec{p}_\mathrm{start}, \mathvec{p}_{\gamma,1}, \mathvec{p}_{\gamma,2}, \ldots, \mathvec{p}_\mathrm{goal} \right) \; ,
\end{equation}
where \(\mathvec{p}_\mathrm{start}\) and \(\mathvec{p}_\mathrm{goal}\) denote the start and goal positions, respectively, and the intermediate positions are given by the intersection of the path with the polygon edges of the traversable regions forming the corridor. 
Although string pulling is performed purely in \(\mathbb{R}^3\) without explicit yaw information, the resulting path remains feasible since it stays within the corridor \(\Pi_\mathrm{compact}\). As the corridor is obtained from an A* search over yaw-specific layers, feasible yaw transitions between consecutive traversed regions are already guaranteed.
For details of the string pulling procedure, we refer to~\cite{mononen2010funnel}.

\subsubsection{Yaw Refinement}

The path \(\gamma\) alters the direction of motion along each path segment, which necessitates replanning the robot's yaw along the path. To this end, we perform yaw refinement constrained to \(\gamma\). For each position along \(\gamma\), we identify the traversable regions containing it and retrieve the corresponding feasible yaw channel set. Vertices are then registered for all feasible yaw angles at these positions, and a new navigation graph \(G\) is constructed following the procedure described in~\Cref{subsubsec:se2_graph}. An A* search on this graph finds an optimal state sequence that is consistent with both the refined straightened path and the yaw-dependent traversability constraints. The cost and heuristic functions used in this stage are identical to those defined in~\Cref{subsubsec:initial_Search}.

\subsection{Online Generation of SE(2) NavMesh}

The online method incrementally generates and updates the SE(2) NavMesh using point cloud measurements of the surrounding environment acquired by onboard sensors. The following description focuses on the components specific to the online method, since most processing steps reuse the procedures introduced in~\Cref{subsec:offline_generation}. 

\subsubsection{Map Update}

We employ Voxblox~\cite{oleynikova2017voxblox} to process the input point clouds and incrementally construct a triangle mesh representation of the environment. Voxblox maintains spatial information in a block-based structure, where space is discretized into cubic Voxblox voxels with an edge length of \(s_\mathrm{map}\). To avoid ambiguity, we use the term \textit{Voxblox voxel} for the Voxblox map representation, while \textit{voxel} refers to the discretization used in the SE(2) NavMesh unless otherwise specified. Voxblox voxels are grouped into cubic blocks of size \(d_\mathrm{block} \times d_\mathrm{block} \times d_\mathrm{block}\), where \(d_\mathrm{block}\) denotes the number of Voxblox voxels along each dimension. The triangle mesh in these blocks is generated from the TSDF values using the marching cubes algorithm~\cite{lorensen1987marching}.

\subsubsection{SE(2) NavMesh Update}

\begin{figure}[t]
    \centering
    \begin{subfigure}{0.48\linewidth}
        \centering
        \includegraphics[width=\linewidth]{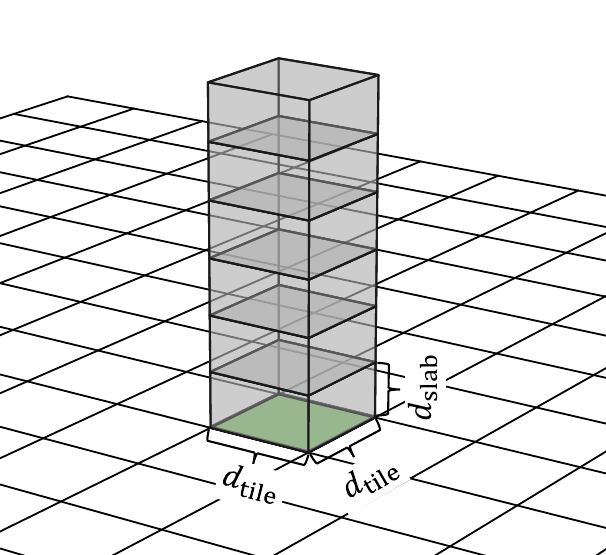}
        \caption{Slabs.}
        \label{fig:tile_slab}      
    \end{subfigure}
    \begin{subfigure}{0.48\linewidth}
        \centering
        \includegraphics[width=\linewidth]{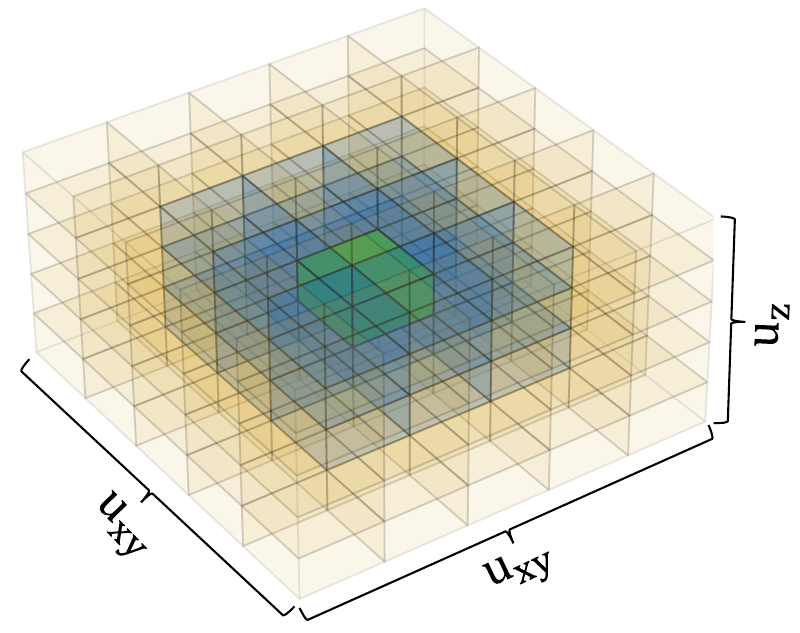}
        \caption{Update window.}
        \label{fig:slab_update}       
    \end{subfigure}

    \caption{Slab-based online SE(2) NavMesh update. (a) A tile is divided along the vertical axis into equal-height slabs, shown as gray cuboids. (b) During an update, the central green slab contains the changed map geometry. The green and blue slabs require local SE(2) NavMesh updates, while geometry from the green, blue, and yellow slabs is used for the updates.}
    \label{fig:se2_navmesh_update}
\end{figure}

Whenever the map is updated, the SE(2) NavMesh must be updated accordingly. Reconstructing the SE(2) NavMesh over the entire map after every update is inefficient, as the map is expanding and map updates are typically spatially localized. We therefore perform local SE(2) NavMesh updates restricted to affected regions of the map. To support efficient local updates, each tile introduced in~\Cref{subsubsec:tile_paritioning} is further divided along the vertical axis into multiple slabs, as illustrated in~\Cref{fig:tile_slab}. Each slab spans \(d_\mathrm{tile} \times d_\mathrm{tile} \times d_\mathrm{slab}\) voxels, where \(d_\mathrm{slab}\) denotes the slab height in voxels. Slab boundaries are aligned with the Voxblox block grid, and each slab dimension is chosen as an integer multiple of the corresponding block dimension to enable efficient indexing and geometry retrieval. When the map geometry within a slab changes, all slabs within a surrounding 3D update window centered at the modified slab are affected and must be updated. The update window spans \(u_\mathrm{xy} \times u_\mathrm{xy} \times u_\mathrm{z}\) slabs, where \(u_\mathrm{xy}\) and \(u_\mathrm{z}\) denote the number of covered slabs in the planar and vertical direction, respectively:
\begin{equation}
    u_\mathrm{xy}  = \left(1 + 2 \left\lceil \frac{d_\mathrm{border}}{d_\mathrm{tile}} \right\rceil \right), 
\end{equation}
\begin{equation}
    u_\mathrm{z} = \left(1 + 2 \left\lceil \frac{d_\mathrm{zborder}}{d_\mathrm{slab}} \right\rceil \right).
\end{equation}
Here, \(d_\mathrm{border}\) is the planar margin in voxels defined in~\Cref{subsubsec:tile_paritioning}, while \(d_\mathrm{zborder}\) is the vertical margin. The latter is chosen as the robot height in voxels plus an additional safety margin of several voxels. For each affected slab, surrounding geometry is retrieved from neighboring slabs within the same spatial window. The local SE(2) NavMesh for each affected slab is then reconstructed by reusing the corresponding steps described in the offline method~(\Cref{subsubsec:method_offline_map_voxelization,subsubsec:method_offline_yaw_feasibility,subsubsec:method_offline_polygon_generation}). The updated local structures are subsequently merged into the global SE(2) NavMesh, where connectivity is re-established with vertically adjacent slabs in the same tile and horizontally adjacent slabs at the same height level in neighboring tiles. \Cref{fig:slab_update} illustrates the spatial relationship among the slab containing the updated map geometry, the affected slabs requiring local SE(2) NavMesh updates, and the surrounding geometry access region. The illustration assumes that the required planar and vertical margins do not exceed the dimensions of a single slab.

\section{Experiments}
\label{sec:experiments}

We conduct extensive experiments to assess the proposed SE(2) NavMesh in terms of representation quality, planning performance, and real-world deployability. First, across multiple simulated environments, we compare SE(2) NavMesh with the classical NavMesh in traversable region coverage. Second, using the generated SE(2) NavMesh, we evaluate the proposed ASA pathfinding pipeline, quantify the contribution of its individual stages, and benchmark it against representative sampling-based planners using success rate, path cost, and planning time. Finally, we deploy the online SE(2) NavMesh system on a real robot to verify real-time online updates and demonstrate autonomous navigation in real-world environments.

\subsection{Comparison of NavMesh and SE(2) NavMesh}
\label{subsec:comparison_navemsh_offline_se2}

\begin{figure*}[t]
    \centering
    \includegraphics[width=0.98\linewidth]{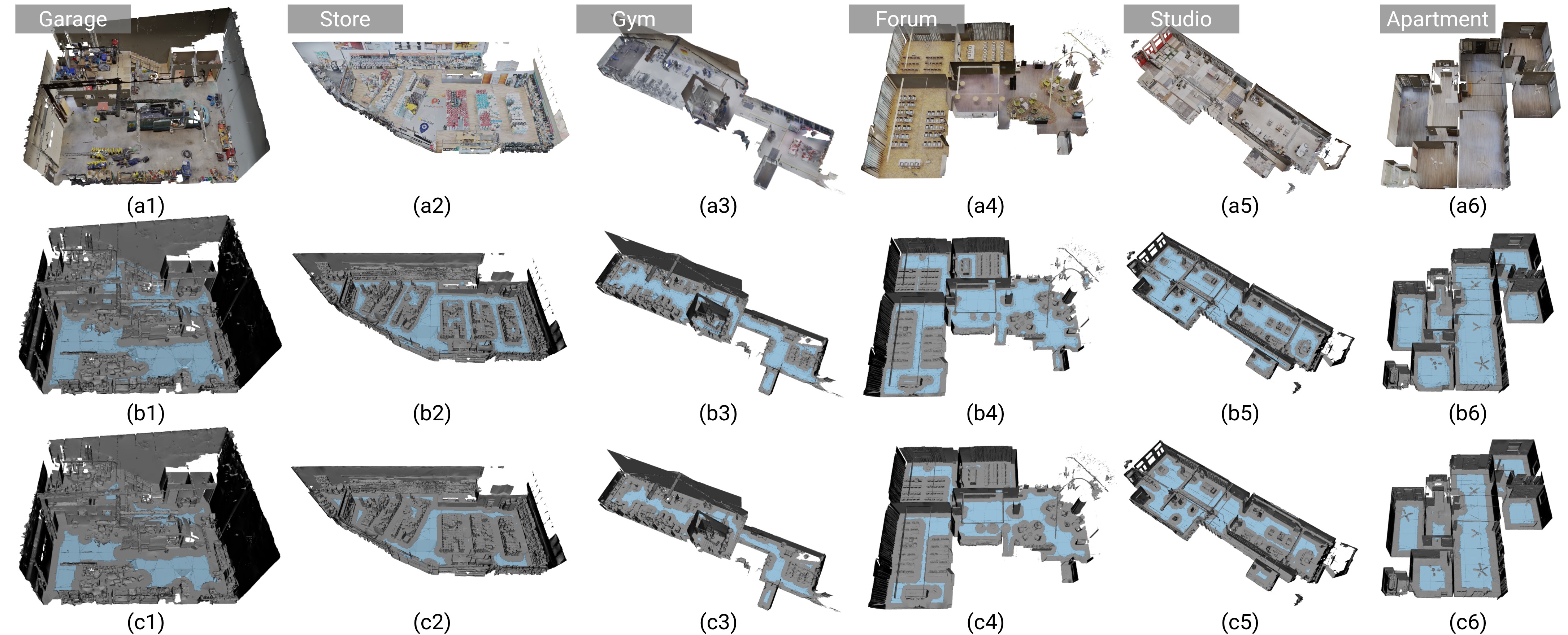}
    \caption{Offline mesh generation results across six simulation scenes. The first row shows the 3D scene models: (a1) \textit{Garage}, (a2) \textit{Store}, (a3) \textit{Gym}, (a4) \textit{Forum}, (a5) \textit{Studio}, and (a6) \textit{Apartment}. The second row (b1-b6) shows traversable regions generated by SE(2) NavMesh, and the third row (c1-c6) shows those generated by the NavMesh. Traversable regions are colored blue.}    
    \label{fig:offline_generation_results}
\end{figure*}

We compare the generation of the NavMesh and the proposed offline SE(2) NavMesh in terms of the traversable area and the generation time. Simulation experiments are conducted in six diverse indoor environments selected from HM3D~\cite{ramakrishnan2021habitat}, as shown in~\Cref{fig:offline_generation_results}.
For each scene, both meshes are constructed using identical robot parameters and mesh generation parameters. The simulated robot is modeled after the ANYmal~\cite{hutter2016anymal}, a quadruped robot capable of locomotion over stairs and sloped terrain. The robot parameters used in the simulation are listed in~\Cref{tab:ANYmal}. Dimensions and traversal capabilities follow~\cite{anymal2022specs}, while the locomotion speed is conservatively chosen for safe navigation. The parameters for mesh generation are provided in~\Cref{tab:sim_mesh_params}.
For the NavMesh, the robot circumradius \(r_\mathrm{circ}\) is derived from the robot length \(l_\mathrm{robot}\) and width \(w_\mathrm{robot}\). All simulation experiments are conducted on an Intel i7-11800H @ \SI{2.30}{\giga\hertz} CPU.

\begin{table}[!t]
    \centering
    \caption{ANYmal Parameters}
    \label{tab:ANYmal}
    \begin{tabular}{@{} l l c @{}}
         \toprule
         Category & Parameters & Values \\
         \midrule
         Dimensions & \(l_\mathrm{robot}, w_\mathrm{robot}, h_\mathrm{robot}\) & \( \SI{0.93}{\meter}, \SI{0.53}{\meter}, \SI{0.89}{\meter}\) \\
         Traversal Cap. & \(h_\mathrm{step}, \theta_\mathrm{climb}\) & \(\SI{0.25}{\meter}, \SI{30}{\degree}\) \\
         Speed & \(v_\mathrm{long}, v_\mathrm{lat}, \omega\) & \(\SI{0.5}{\meter / \second}, \SI{0.1}{\meter / \second}, \SI{0.5}{\radian / \second}\) \\
         \bottomrule         
    \end{tabular}
\end{table}

\begin{table}[!t]
    \centering
    \caption{Mesh Generation Parameters}
    \label{tab:sim_mesh_params}
    \begin{tabular}{@{} l l c @{}} 
        \toprule
        Category & Parameters & Values \\
        \midrule
        Voxel & \( s_{\mathrm{voxel}} \times s_{\mathrm{voxel}} \times h_{\mathrm{voxel}} \) & \(\SI{0.1}{\meter} \times \SI{0.1}{\meter} \times \SI{0.1}{\meter} \) \\
        Tile  & \( d_{\mathrm{tile}} \times  d_{\mathrm{tile}}\) & \(16 \times 16\) \\
        Yaw   & \( N_{\Psi} \) & 40 \\
        \bottomrule
    \end{tabular}
\end{table}

The selected scenes cover diverse layouts and navigation challenges, including multi-level structures, narrow passages, and doorways.
\begin{itemize}
    \item \textit{Garage}: A two-level environment with a loft. The ground floor contains many objects and is connected to the loft by a narrow staircase. Its dimensions are \(\SI{15.7}{\meter} \times \SI{12.5}{\meter} \times \SI{8.2}{\meter} \).
    \item \textit{Store}: A single-level environment containing many shelves. Narrow aisles are formed between the shelves. The overall scene dimensions are \(\SI{30.3}{\meter} \times \SI{13.0}{\meter} \times \SI{5.1}{\meter}\).
    \item \textit{Gym}: A two-level environment in which the upper level contains fitness equipment. The two levels are connected via a staircase. Its dimensions are \(\SI{40.4}{\meter} \times \SI{11.7}{\meter} \times \SI{11.3}{\meter}\).
    \item \textit{Forum}: A single-level environment consisting of several meeting rooms. Each meeting room is connected to the main hall through a door. Its dimensions are \(\SI{31.4}{\meter} \times \SI{30.0}{\meter} \times \SI{4.7}{\meter}\).
    \item \textit{Studio}: A single-level environment with sparse furnishings. The main area and a secondary area are connected, with a narrow corridor leading to a restroom. Its dimensions are \(\SI{39.4}{\meter} \times \SI{12.2}{\meter} \times \SI{4.2}{\meter}\).
    \item \textit{Apartment}: A single-level environment representing an empty apartment composed of multiple rooms. Its dimensions are \(\SI{16.6}{\meter} \times \SI{15.1}{\meter} \times \SI{3.0}{\meter}\).
\end{itemize}

\begin{figure}[t]
    \centering
    \begin{subfigure}{0.32\linewidth}
        \centering
        \includegraphics[width=\linewidth]{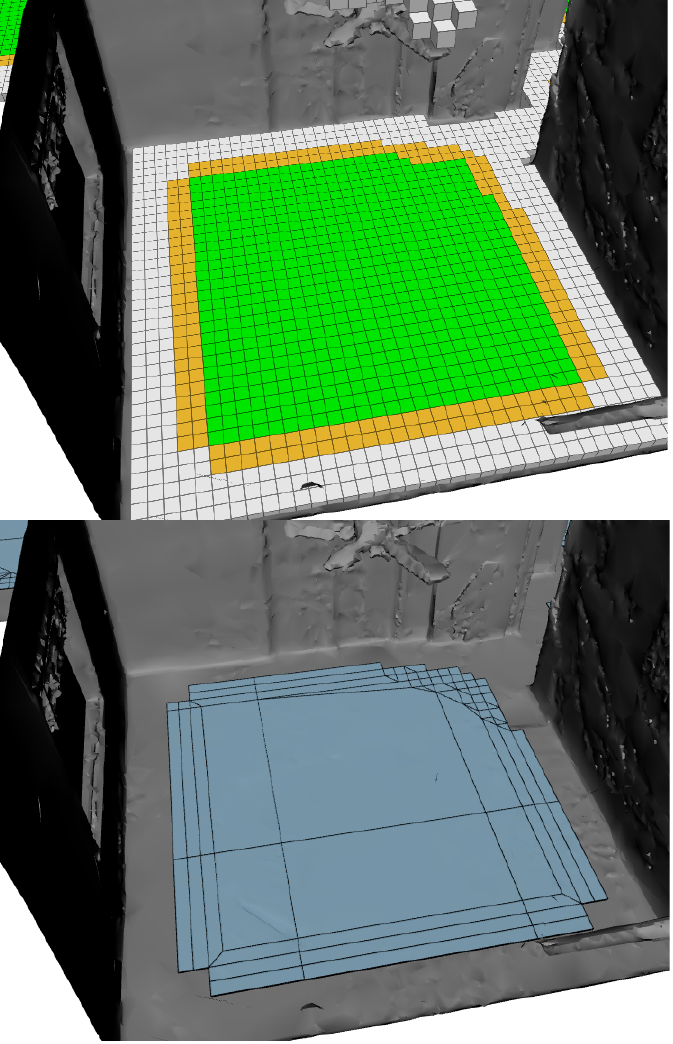}
        \caption{Empty room.}
        \label{fig:SE(2)NavMesh_closeup_room}      
    \end{subfigure}
    \begin{subfigure}{0.32\linewidth}
        \centering
        \includegraphics[width=\linewidth]{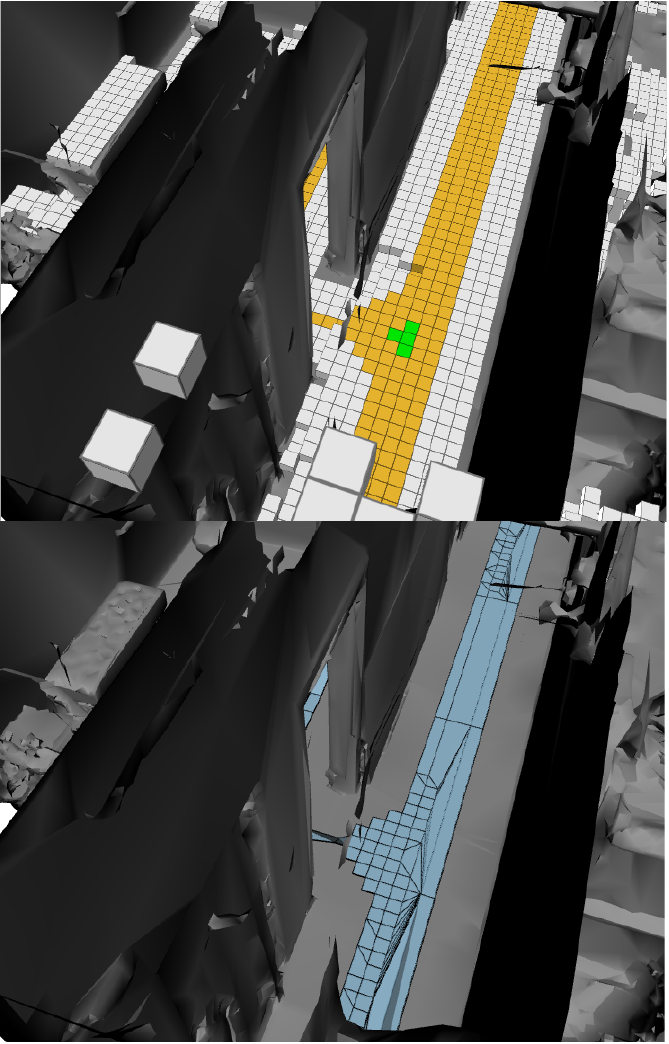}
        \caption{Narrow corridor.}
        \label{fig:SE(2)NavMesh_closeup_corridor}       
    \end{subfigure}
    \begin{subfigure}{0.3264\linewidth}
        \centering
        \includegraphics[width=\linewidth]{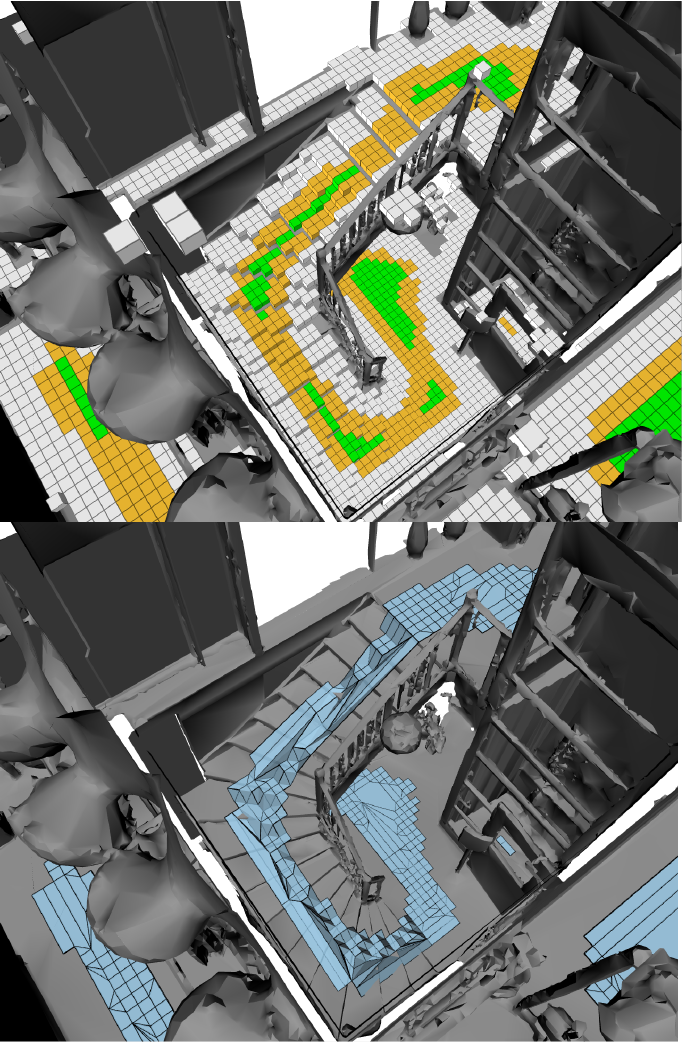}
        \caption{Staircase.}
        \label{fig:SE(2)NavMesh_closeup_staircase}       
    \end{subfigure}
    
    \caption{Close-up views of voxel classification (top row) and generated SE(2) NavMesh polygons (bottom row) of an empty room in \textit{Apartment}, a narrow corridor in \textit{Studio}, and a staircase in \textit{Gym}. Safe voxels are shown in green, and restricted voxels are shown in yellow.}
    \label{fig:SE(2)NavMesh_closeup}
\end{figure}

\Cref{fig:offline_generation_results} compares the traversable regions generated by SE(2) NavMesh and the NavMesh. By accounting for yaw-dependent traversability, the proposed method preserves narrow passages and near-boundary regions that are discarded by the NavMesh. \Cref{fig:SE(2)NavMesh_closeup} provides close-up views of voxel classification and the resulting traversable region polygon generated by SE(2) NavMesh. In open areas (\Cref{fig:SE(2)NavMesh_closeup_room}), safe voxels can be merged into large convex polygons (safe traversable regions). In contrast, in constrained environments such as corridors (\Cref{fig:SE(2)NavMesh_closeup_corridor}) and staircases (\Cref{fig:SE(2)NavMesh_closeup_staircase}), the feasible yaw channel set varies frequently across neighboring voxels, preventing merging and resulting in a greater number of smaller polygons that encode local yaw-dependent traversability. Together, this property naturally yields an adaptive spatial discretization, allowing efficient traversal through safe regions while maintaining fine planning resolution in restricted regions. 

\begin{figure*}[t]
    \centering
    \includegraphics[width = 0.9\textwidth]{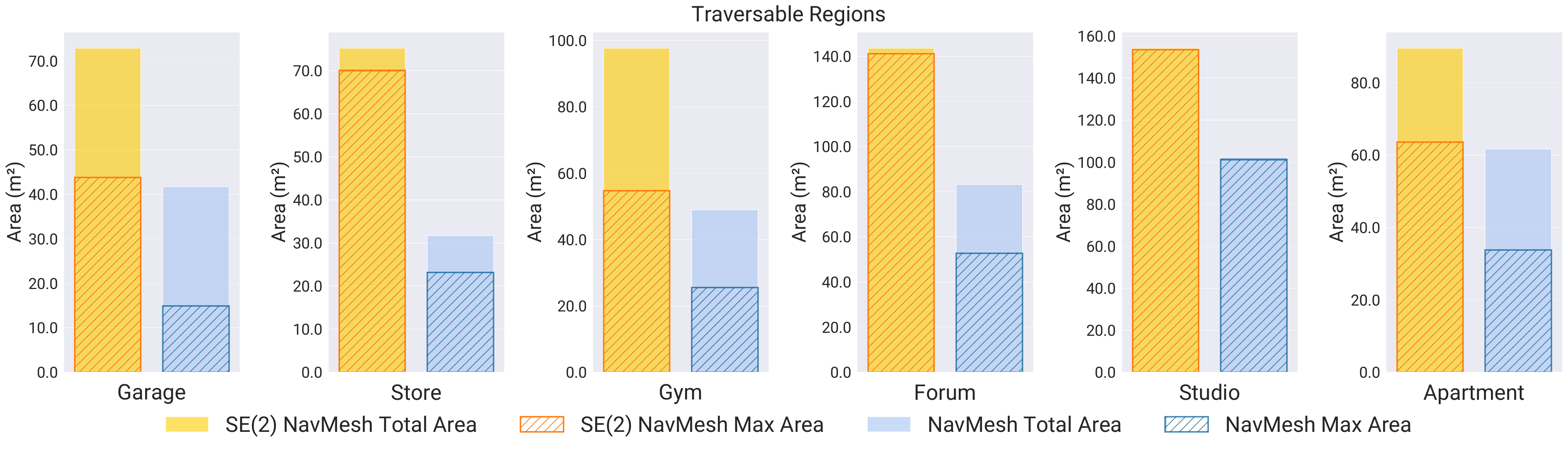}
    \caption{Comparison of traversable regions generated by the NavMesh and SE(2) NavMesh across simulation scenes.}
    \label{fig:traversable_regions}
\end{figure*}

A comparison of the traversable regions generated by the NavMesh and the proposed SE(2) NavMesh is shown in~\Cref{fig:traversable_regions}. For each scene, we evaluate both the total area of the traversable regions and the area of the largest connected traversable component. Owing to the inclusion of restricted traversable regions and a more accurate evaluation of yaw-dependent traversability, SE(2) NavMesh consistently outperforms the NavMesh across all scenes, capturing over \SI{50}{\percent} more traversable area in every evaluated scene. Furthermore, restricted traversable regions bridge previously disconnected safe traversable regions, leading to a substantial increase in the size of the largest connected traversable component.

The time consumed by different stages of the two generation pipelines across all simulation scenes is summarized in~\Cref{tab:offline_generation_time}. For SE(2) NavMesh, the total build time consists of continuous-yaw footprint mask generation and mesh construction. The mask generation stage introduces only a small overhead, averaging approximately \SI{2.35}{\milli\second}. Despite producing substantially more region polygons than the NavMesh (approximately 8 to 10 times more), SE(2) NavMesh requires only 2 to 3 times longer total build time. This indicates that the proposed method greatly enriches the representation of traversable regions at a modest additional computational cost.

\begin{table}[!t]
    \centering
    \caption{Build Time and Region-Polygon Count for the NavMesh and SE(2) NavMesh across simulation scenes}
    \label{tab:offline_generation_time}
    \begin{tabular}{@{} l c c c c @{}}
        \toprule
        Method & \thead{Mask Gen.\\(ms)} & \thead{Mesh Const.\\(ms)} & \thead{Total\\(ms)} & \thead{\# Region\\Polygons}\\
        \midrule
        \multicolumn{5}{@{}l}{\textit{Garage}}\\
        \quad NavMesh       & -    & 37.80 & 37.80 & 278\\
        \quad SE(2) NavMesh & 2.44 & 86.25 & 88.69 & 2859\\
        \midrule
        \multicolumn{5}{@{}l}{\textit{Store}}\\
        \quad NavMesh       & -    & 48.15 & 48.15 & 374\\
        \quad SE(2) NavMesh & 2.48 & 91.96 & 94.44 & 4007\\
        \midrule
        \multicolumn{5}{@{}l}{\textit{Gym}}\\
        \quad NavMesh       & -    & 43.53 & 43.53 & 478\\
        \quad SE(2) NavMesh & 2.37 & 89.15 & 91.52 & 4362\\
        \midrule
        \multicolumn{5}{@{}l}{\textit{Forum}}\\
        \quad NavMesh       & -    & 42.02 & 42.02 & 628\\
        \quad SE(2) NavMesh & 2.16 & 107.25 & 109.41 & 5055\\
        \midrule
        \multicolumn{5}{@{}l}{\textit{Studio}}\\
        \quad NavMesh       & -    & 42.87 & 42.87 & 603\\
        \quad SE(2) NavMesh & 2.37 & 102.17 & 104.54 & 4521\\
        \midrule
        \multicolumn{5}{@{}l}{\textit{Apartment}}\\
        \quad NavMesh       & -    & 25.83 & 25.83 & 193\\
        \quad SE(2) NavMesh & 2.28 & 60.47 & 62.75 & 1729\\
        \bottomrule
    \end{tabular}
\end{table}

\subsection{Pathfinding Benchmark}

We evaluate pathfinding performance on the SE(2) NavMesh using the proposed ASA strategy and compare it against sampling-based planners, including RRT~\cite{lavalle2001rapidly}, RRT*~\cite{karaman2011sampling}, and PRM~\cite{kavraki2002probabilistic}. All sampling-based planners are implemented using OMPL~\cite{sucan2012open}. The robot parameters and mesh generation parameters are identical to those used in the~\Cref{subsec:comparison_navemsh_offline_se2}. We first describe the configuration of the sampling-based planners, including the state space and the state validity check methods. We then consider two benchmark settings: time-dependent planning performance and randomly sampled start-goal pairs. Finally, we analyze path quality comparisons across its individual stages.

\subsubsection{Sampling-Based Planner Configuration}

\begin{algorithm}[h]
    \caption{Voxel-Based State Validity Checker}
    \label{alg:checker_voxel}
    \begin{algorithmic}[1]
        \renewcommand{\algorithmicrequire}{\textbf{Input:}}
        \renewcommand{\algorithmicensure}{\textbf{Output:}}

        \Require State \(\mathvec{s}\), global terrain voxel map \(\mathset{W}_\mathrm{global}\), height tolerance \(\varepsilon_z\)
        \Ensure Valid \(\in \{\text{true}, \text{false}\}\)

        \State \(v \gets \textsc{FindNearestVoxelInColumn} (\mathset{W}_\mathrm{global}, \mathvec{s}) \)
        
        \If{\(v = \text{null}\)}
            \State \Return false
        \EndIf

        \State \(z \gets \textsc{GetVoxelHeight}(v)\)

        \If{\(|z_s-z| > \varepsilon_z\)}
            \State \Return false
        \EndIf

        \State \(M \gets \textsc{GenerateFootprintMaskFromYaw}(\psi_s)\)
        
        \State \textsc{AlignReferenceCell}\((M, v)\)
        
        \If{not \textsc{MaskFeasible}\( (M,v,\mathset{W}_\mathrm{global}) \)}
            \State \Return false
        \EndIf

        \State \Return true
    \end{algorithmic}
\end{algorithm}

The sampling-based planners sample states \(\mathvec{s} = (\mathvec{p}_s, \psi_s) \in \mathbb{R}^3 \times S^1\), where \(\mathvec{p}_s = (x_s,y_s,z_s)\). Each sampled state is evaluated by a state validity checker, which verifies both individual states and interpolated states along candidate edges. We implement two state validity checkers.
The first checker uses a global terrain voxel generated following the procedure described in~\Cref{subsubsec:method_offline_map_voxelization}, where each voxel is annotated as walkable or non-walkable. The sampled position is projected onto the nearest terrain voxel along the corresponding vertical column and is accepted only if the vertical deviation is below a predefined threshold. If the voxel is walkable, an isolated-yaw footprint mask corresponding to \(\psi_s\) is further evaluated, and the state is valid when all covered voxels are walkable.
The second checker uses the generated SE(2) NavMesh: the sampled position must lie within a traversable region and satisfy the same vertical threshold, and the state is valid only if its yaw \(\psi_s\) lies within that region's feasible yaw range. The workflows of the two checkers are presented in~\Cref{alg:checker_voxel,alg:checker_se2}.
Both checkers are integrated with RRT, RRT*, and PRM. Planners using the voxel-based checker retain their original names, while those using the SE(2) NavMesh-based checker are referred to as RRT-SE2NM, RRT*-SE2NM, and PRM-SE2NM, respectively.

\begin{algorithm}[t]
    \caption{SE(2) NavMesh-Based State Validity Checker}
    \label{alg:checker_se2}
    \begin{algorithmic}[1]
        \renewcommand{\algorithmicrequire}{\textbf{Input:}}
        \renewcommand{\algorithmicensure}{\textbf{Output:}}

        \Require State \(\mathvec{s}\),  SE(2) NavMesh \(\mathset{M}_\mathrm{nav}\), height tolerance \(\varepsilon_z\)
        \Ensure Valid \(\in \{\text{true}, \text{false}\}\)

        \State \(\mathset{R} \gets \textsc{FindNearestRegion}(\mathset{M}_\mathrm{nav}, \mathvec{s})\)

        \If{\(\mathset{R} = \text{null}\)}
            \State \Return false
        \EndIf

        \State \(z \gets \textsc{GetReferenceHeight}(\mathset{R},\mathvec{s})\)

        \If{\(|z_s-z| > \varepsilon_z\)}
            \State \Return false
        \EndIf

        \If{ not \(\textsc{IsYawFeasible}(\mathset{R}, \psi_s)\)}
            \State \Return false 
        \EndIf
        
        \State \Return true
    \end{algorithmic}
\end{algorithm}

Interpolated states partition the motion between two connected states into multiple segments. Within each segment, the robot is assumed to traverse the path with the yaw of the starting state and rotate only at the segment end. The total motion cost is obtained by summing the segment costs computed using the formulation in~\Cref{subsec:method_pathfinding}.

\subsubsection{Time-Dependent Planning Performance}

\begin{figure*}[t]
    \centering
    \includegraphics[width=0.95\linewidth]{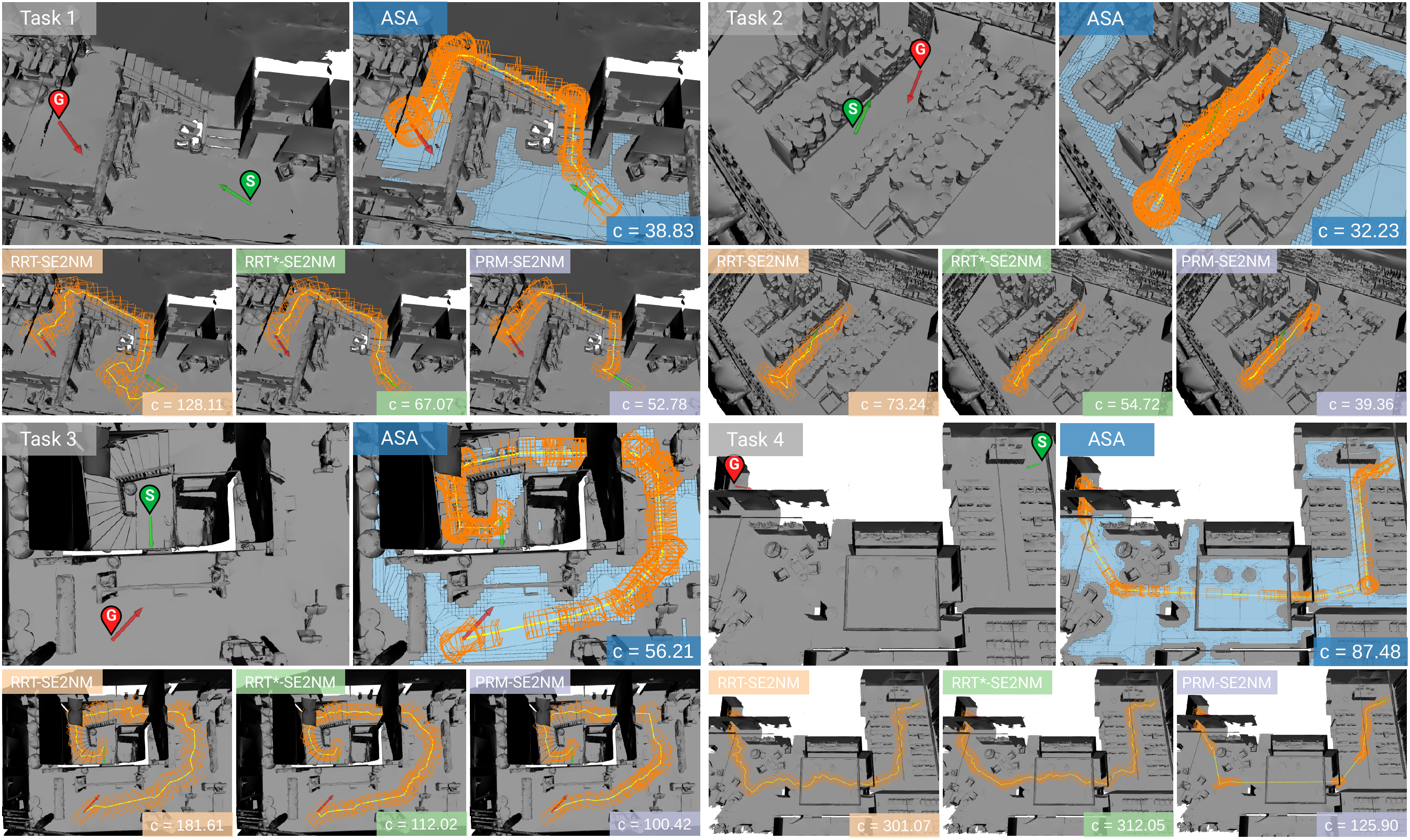}

    \caption{Pathfinding tasks and example solutions in constrained environments. For each task, the left image in the first row shows the environment and task setup. Green and red arrows denote the start and goal states. The planned path is visualized as a yellow polyline, with orange wireframe boxes indicating the robot states. The path cost \(c\) is shown in the bottom right corner of each example solution figure. Compared with sampling-based methods, ASA produces straighter paths with smoother rotation.}
    \label{fig:performance_over_time_task1_4_vis}
\end{figure*}

\begin{figure*}[!t]
    \centering
    \includegraphics[width = 0.95\linewidth]{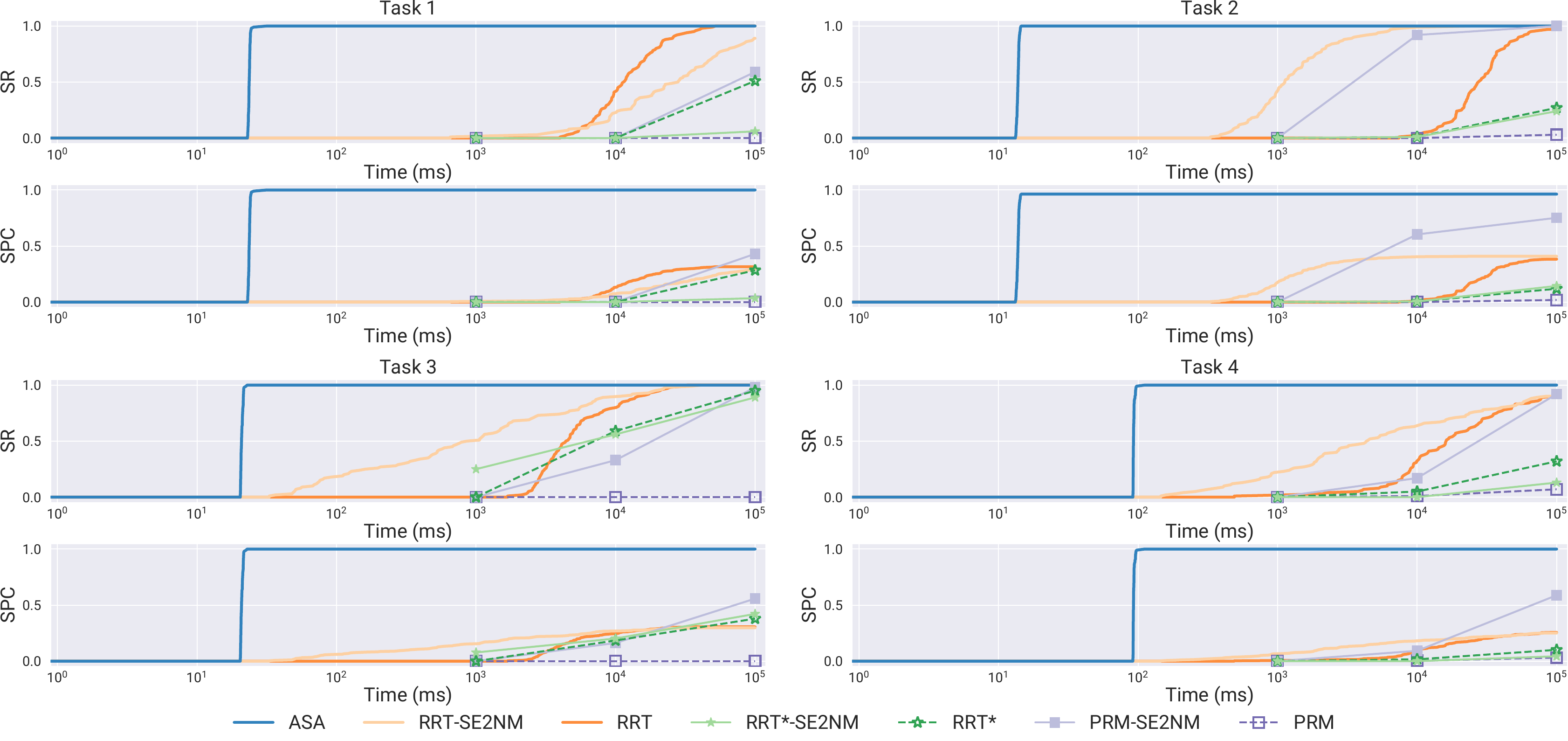}
    
    \caption{Trends of SR and SPC over planning time for different planners in constrained environments. ASA achieves the best overall performance, while PRM-SE2NM generally achieves the highest SPC among sampling-based planners.}
    \label{fig:performance_over_time_task1_4_plt}
\end{figure*}

\begin{figure}[!t]
    \begin{subfigure}{\linewidth}
        \centering
        \includegraphics[width=0.98\linewidth]{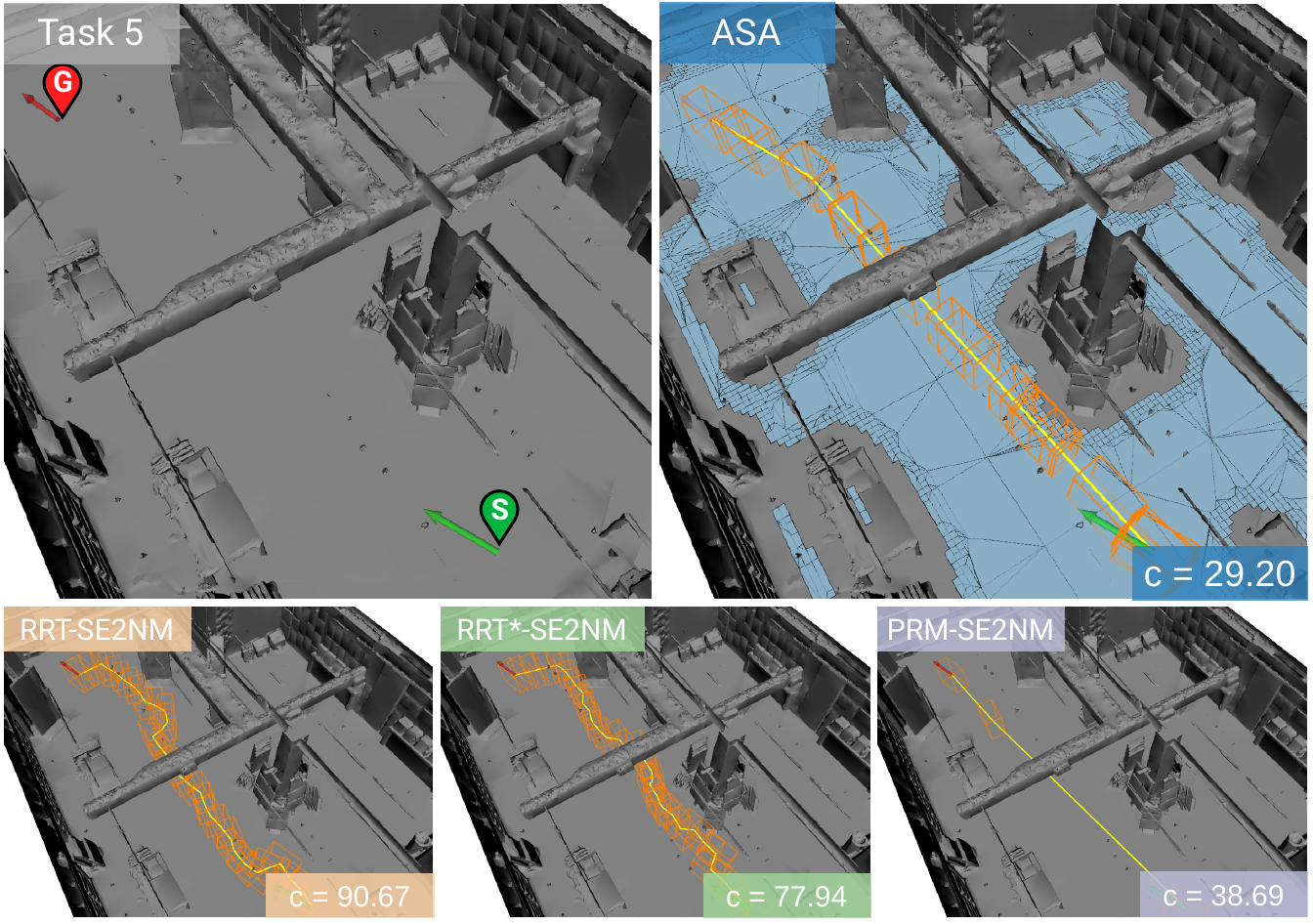}
        \caption{Task setup and example solutions. }
        \label{fig:benchmark_task5_vis}
    \end{subfigure}
    
    \begin{subfigure}{\linewidth}
        \centering
        \includegraphics[width=0.95\linewidth]{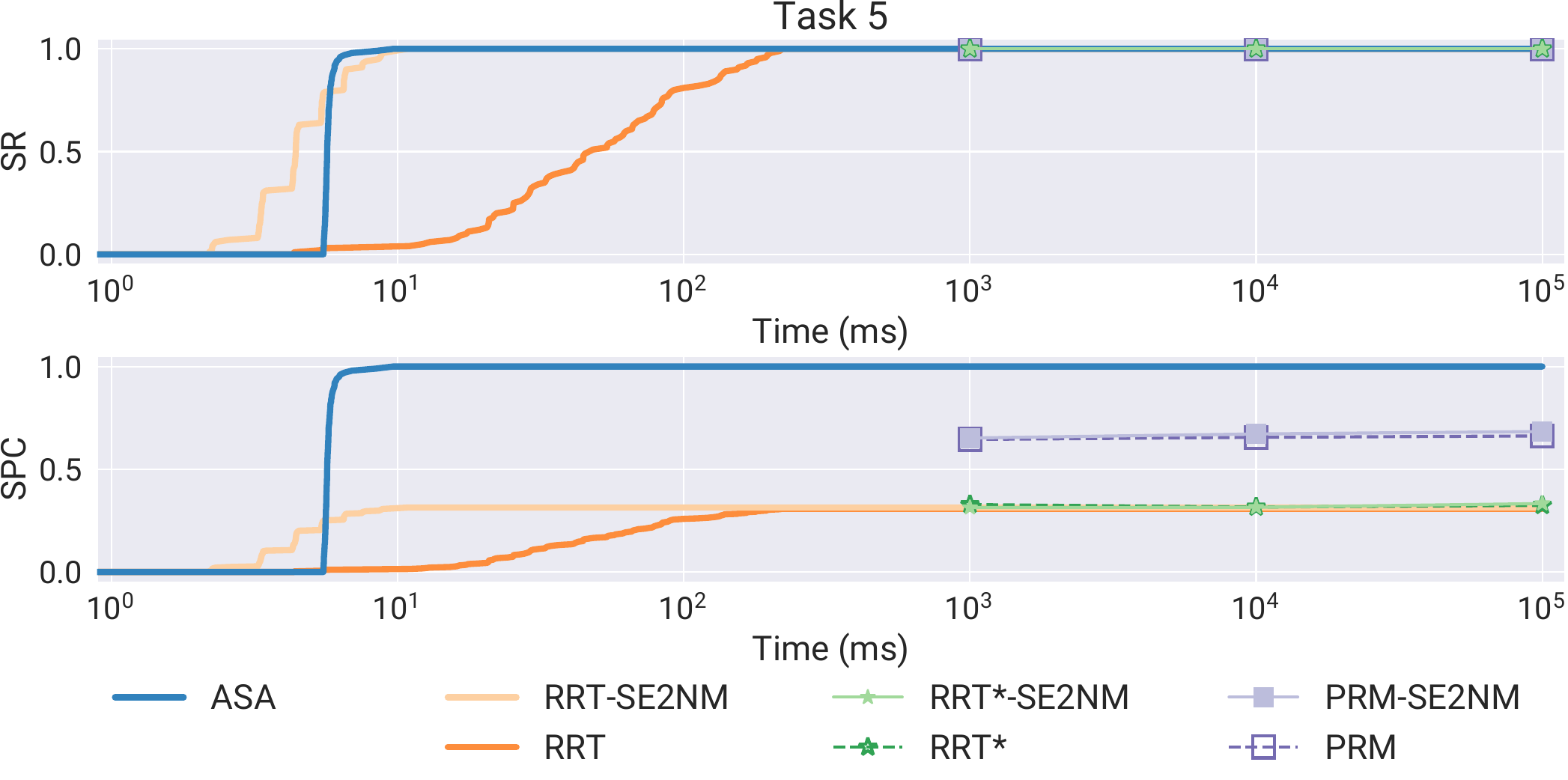}
        \caption{Trends of SR and SPC over planning time for different planners.}
        \label{fig:benchmark_test5_plt}
    \end{subfigure}

    \caption{Pathfinding performance in Task 5, an open-space scene. RRT-SE2NM requires slightly less planning time in open space, while ASA consistently produces lower-cost paths.}
    \label{fig:performance_over_time_task5}
    
\end{figure}

We design five pathfinding tasks to evaluate the time-dependent performance of ASA and the sampling-based planners. For each task, the start and goal states are selected within the same connected traversable component to ensure that a feasible path exists. This allows the comparison to focus on planning efficiency and solution quality rather than reachability. For RRT and RRT-SE2NM, the maximum planning time is fixed at \SI{100}{\second}. RRT*, PRM, RRT*-SE2NM, and PRM-SE2NM are asymptotically optimal planners whose solution quality improves as planning time increases. Therefore, these planners are evaluated under planning time limits of \SI{1}{\second}, \SI{10}{\second}, and \SI{100}{\second} to analyze their performance over time. For each task, every method is executed for 100 trials. 

We evaluate pathfinding performance using two metrics: the success rate (SR) and the success weighted by inverse path cost (SPC). We define SPC as a success-weighted normalized efficiency metric, inspired by the success weighted by inverse path length (SPL)~\cite{anderson2018evaluation} and the success weighted by completion time (SCT)~\cite{yokoyama2021success}, given by
\begin{equation}
    \mathrm{SPC} = \frac{1}{N}\sum_{i=1}^{N} S_i \frac{c_i^*}{\max(c_i,c_i^*)} \; .
\end{equation}
Here, \(N\) denotes the total number of pathfinding problems. \(S_i\) is a binary indicator of success for the evaluated method on the \(i\)-th problem. \(c_i\) denotes the path cost produced by that method, and \(c_i^*\) is the minimum path cost achieved for the problem, which in this experiment is taken as the minimum path cost obtained by all methods across all trials. This metric jointly accounts for both SR and solution quality in terms of path cost. 

\Cref{fig:performance_over_time_task1_4_vis} depicts Tasks 1 to 4, in which the start and goal are separated by challenging structures such as stairs (Task 1 and Task 3), narrow passages (Task 2), and narrow doorways (Task 3 and Task 4). These features make the tasks representative of geometrically constrained pathfinding scenarios. The pathfinding performance of all methods in Tasks 1 to 4 is presented in~\Cref{fig:performance_over_time_task1_4_plt}. In addition, Task 5 (shown in~\Cref{fig:benchmark_task5_vis}) represents a contrast case in which the environment between the start and goal is relatively open and contains no narrow regions.

Across Tasks 1 to 4, ASA consistently requires the shortest planning time while achieving the highest SPC. This demonstrates that the proposed strategy, together with the generated graph, enables rapid search in constrained environments and obtains low-cost paths. Among the sampling-based planners, PRM-SE2NM generally achieves the highest SPC. However, it requires substantially more planning time to construct a sufficiently dense roadmap and optimize the path before reaching comparable solution quality. In Task 5, where the environment is relatively open, all methods achieve a SR of \SI{100}{\percent} within the given time limits, as shown in~\Cref{fig:benchmark_test5_plt}. In this setting, RRT-SE2NM achieves slightly shorter planning times than ASA. This is because the sampling-based exploration of RRT-SE2NM can rapidly connect the start and goal in open space without the need to search over the mesh structure. However, ASA produces lower-cost paths, as its graph-based search enables more cost-efficient pathfinding.

Across the sampling-based planner experiments, the SE(2) NavMesh-based state validity checker achieves an average per-state checking time on the order of \qty{0.1}{\micro\second}, which is significantly faster than the voxel-based state validity checker, whose average per-state checking time is on the order of \qty{10}{\micro\second}. This computational advantage allows PRM-SE2NM to sample and validate substantially more states, resulting in noticeably higher SR and SPC compared with PRM. However, faster state validity checking does not necessarily translate into higher SR for all planners. The continuous-yaw footprint mask used in SE(2) NavMesh covers a slightly larger area than the isolated-yaw footprint mask. Consequently, this more conservative design filters out some states that lie too close to obstacles for safety. In particularly narrow regions, such as the staircase entrance in Task 1, this effect makes it more difficult for RRT-SE2NM and RRT*-SE2NM to sample valid states and expand the search tree, leading to slower planning performance than their voxel-based counterparts.

\subsubsection{Randomly Sampled Start-Goal Evaluation}
\label{subsubsec:randomly_sampled_start_goal}

Among the multi-level scenes, \textit{Gym} has the largest SE(2) NavMesh coverage and contains two major traversable components, with the largest traversable components spanning two floors connected by a staircase. Among the single-level scenes, \textit{Studio} provides the largest coverage and is largely connected. These two representative scenes are selected for the randomly sampled start-goal evaluation. In each scene, we randomly sample 100 start-goal pairs. Both the start and goal correspond to valid states within their respective traversable regions. Note that due to the random nature of the sampling, the start and goal do not necessarily belong to the same connected traversable component. For all sampling-based planners, the planning time limit is set to \SI{100}{\second}. The \(c_i^*\) used in SPC is the lowest cost achieved among all methods within the \SI{100}{\second} planning time limit.

\begin{table}[t]
\centering
{
    \sisetup{detect-all}
    \caption{SR and SPC on Randomly Sampled Tasks}
    \label{tab:random_sampling}
    \begin{tabular}{l cc cc} 
        \toprule
        \multirow{2}{*}{\textbf{Planner}} & \multicolumn{2}{c}{\textbf{Gym}} & \multicolumn{2}{c}{\textbf{Studio}} \\
        \cmidrule(lr){2-3} \cmidrule(lr){4-5}
        & SR $\uparrow$ & SPC $\uparrow$ & SR $\uparrow$ & SPC $\uparrow$ \\
        \midrule
        ASA  & \textbf{\SI{41}{\percent}} & \textbf{0.41} & \textbf{\SI{98}{\percent}} & \textbf{0.98} \\
        RRT-SE2NM      & \SI{40}{\percent}        & 0.13          & \SI{97}{\percent}           & 0.29 \\
        RRT*-SE2NM     & \SI{32}{\percent}         & 0.18          & \SI{82}{\percent}          & 0.26 \\
        PRM-SE2NM      & \textbf{\SI{41}{\percent}} & 0.27          & \SI{96}{\percent}          & 0.63 \\
        RRT            & \textbf{\SI{41}{\percent}} & 0.13          & \SI{97}{\percent}          & 0.29 \\
        RRT*           & \SI{36}{\percent}           & 0.16          & \SI{80}{\percent}          & 0.25 \\
        PRM            & \SI{36}{\percent}           & 0.15          & \SI{84}{\percent}          & 0.42 \\
        \bottomrule
    \end{tabular}
}
\end{table}

\Cref{tab:random_sampling} reports the SR and the SPC achieved by ASA and the sampling-based planners in both scenes. The connectivity of traversable regions strongly affects SR: the fragmented traversability structure in \textit{Gym} leads to lower SR across all methods, with an average of \SI{38}{\percent}, while the largely connected \textit{Studio} scene yields substantially higher SR, with an average of \SI{91}{\percent}.
Across both scenes, ASA consistently performs better than all sampling-based methods in overall performance, achieving the highest SR and SPC. For every successful start-goal problem, ASA achieves an average SPC close to 1, indicating that its successful solutions are consistently among the lowest-cost paths found by all methods. While RRT-SE2NM and RRT achieve comparable SR to ASA, their substantially lower SPC values suggest that they often produce higher-cost paths.
Under the current planning time limit, whether state validity checking is performed on the generated SE(2) NavMesh has little effect on RRT and RRT*. In contrast, PRM-SE2NM substantially outperforms PRM, improving the average SR by 8.5 percentage points and increasing the average SPC from 0.29 to 0.45. When operating on the SE(2) NavMesh, state validity can be determined directly by querying the feasible yaw channel set of the corresponding traversable region without generating footprint masks. This allows PRM-SE2NM to perform significantly more sampling within the same time budget, greatly increasing the probability of finding feasible paths. The denser roadmap constructed from these additional samples also improves path quality, resulting in the highest SPC among all sampling-based planners.

\subsubsection{Effectiveness of ASA Pathfinding}

\begin{figure}
    \centering
    \begin{subfigure}{0.5\linewidth}
        \centering
        \includegraphics[width=\linewidth]{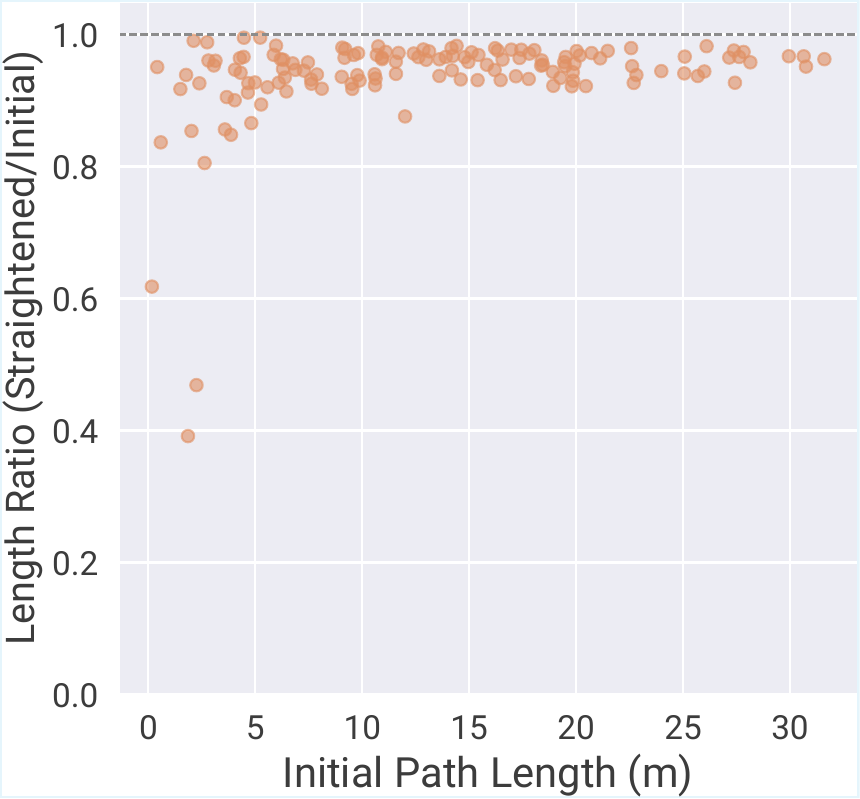}
        \caption{Path length reduction.}
        \label{fig:asa_path_length}
    \end{subfigure}
    \begin{subfigure}{0.48\linewidth}
        \centering
        \includegraphics[width=\linewidth]{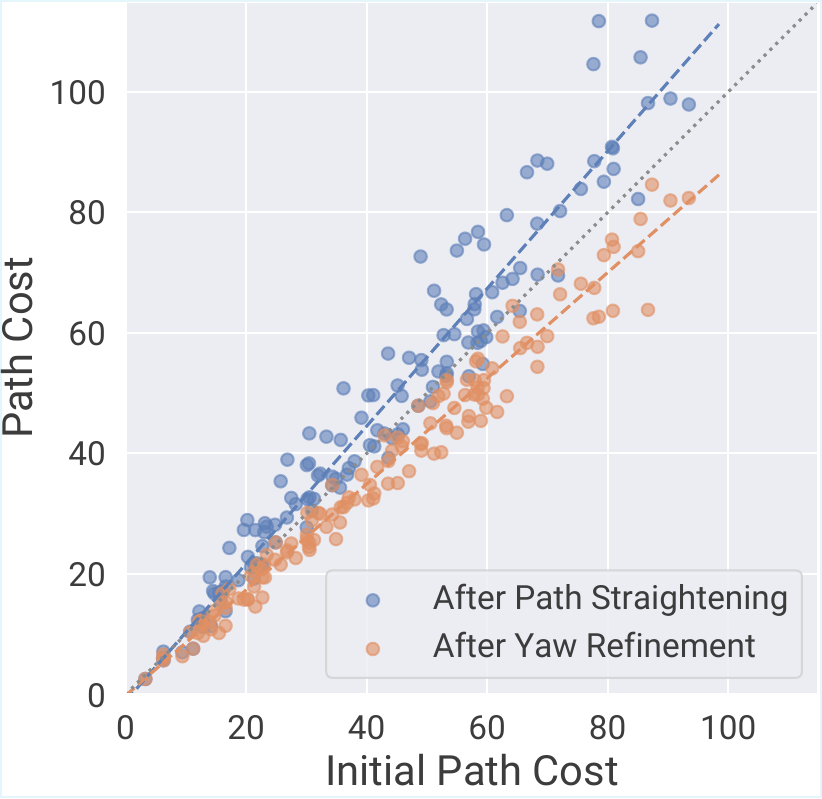}
        \caption{Path cost comparison.}
        \label{fig:asa_path_cost}
    \end{subfigure}
    \caption{Path quality analysis of the ASA pipeline. Path straightening reduces path length but may increase path cost due to mismatched yaw assignments. The yaw refinement stage restores yaw consistency and further reduces the overall path cost.}
\end{figure}

To assess the effectiveness of the ASA pathfinding strategy, we compare path costs and path lengths across its three stages using the same randomly sampled start-goal pairs (\Cref{subsubsec:randomly_sampled_start_goal}). As shown in~\Cref{fig:asa_path_length}, path straightening consistently shortens the paths produced by the initial search, reducing the path length by \SI{6.2}{\percent} on average. However, the refined intermediate positions become mismatched with the yaw assignments obtained in the initial stage, which increases the path cost by approximately \SI{10}{\percent}, as indicated by the blue markers in~\Cref{fig:asa_path_cost}. The final yaw-refinement stage effectively reduces the path cost by re-optimizing yaw at these positions, as shown by the orange markers in~\Cref{fig:asa_path_cost}, and in most cases achieves a lower cost than the initial paths. Overall, the final cost averages \SI{87}{\percent} of the initial path cost.

\subsection{Real-World Experiments}

\begin{figure*}[t]
    \centering
    \includegraphics[width=0.85\linewidth]{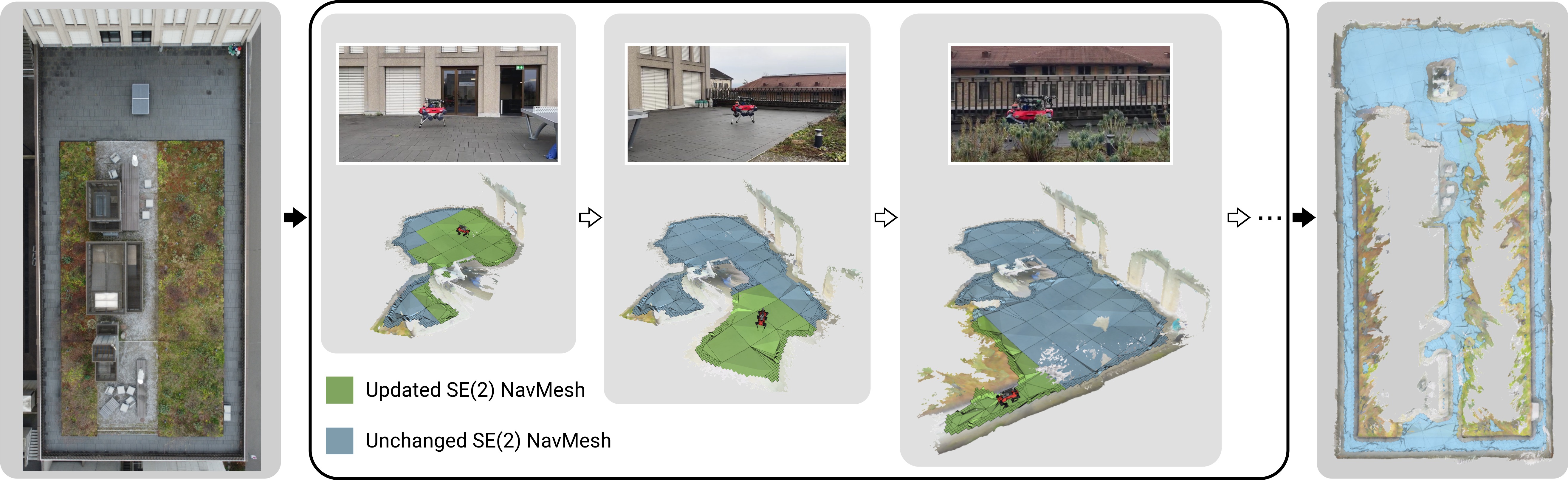}
    \caption{Online SE(2) NavMesh generation in a garden. The left panel shows the real-world scene. The middle panel shows incremental geometry reconstruction and local SE(2) NavMesh updates. Only slabs affected by geometry changes have their local SE(2) NavMesh updated (green), while unaffected slabs retain their existing SE(2) NavMesh (blue). The right panel shows the final reconstructed geometry and the generated SE(2) NavMesh.}
    \label{fig:online_update_experiment}
\end{figure*}

\begin{figure*}[!t]
    \centering
    \includegraphics[width=0.9\linewidth]{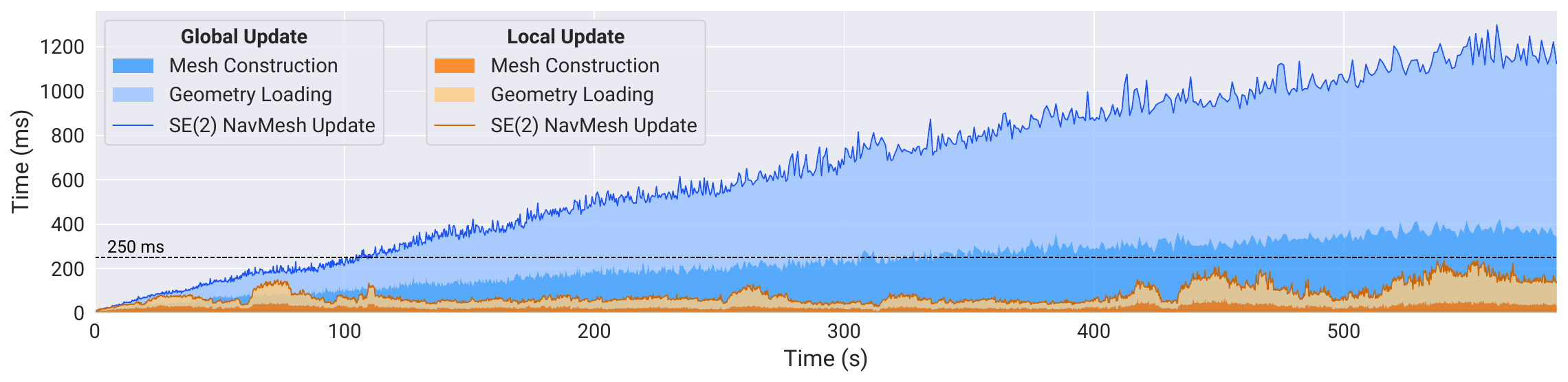}
    \caption{Comparison of local and global SE(2) NavMesh update times during online generation. Local updates maintain the target real-time update rate, whereas global updates incur increasing latency as the reconstructed map grows.}
    \vspace{-0.5em}
    \label{fig:online_update_comparison}
\end{figure*}

\begin{figure*}[t]
    \centering
    \includegraphics[width=0.75\linewidth]{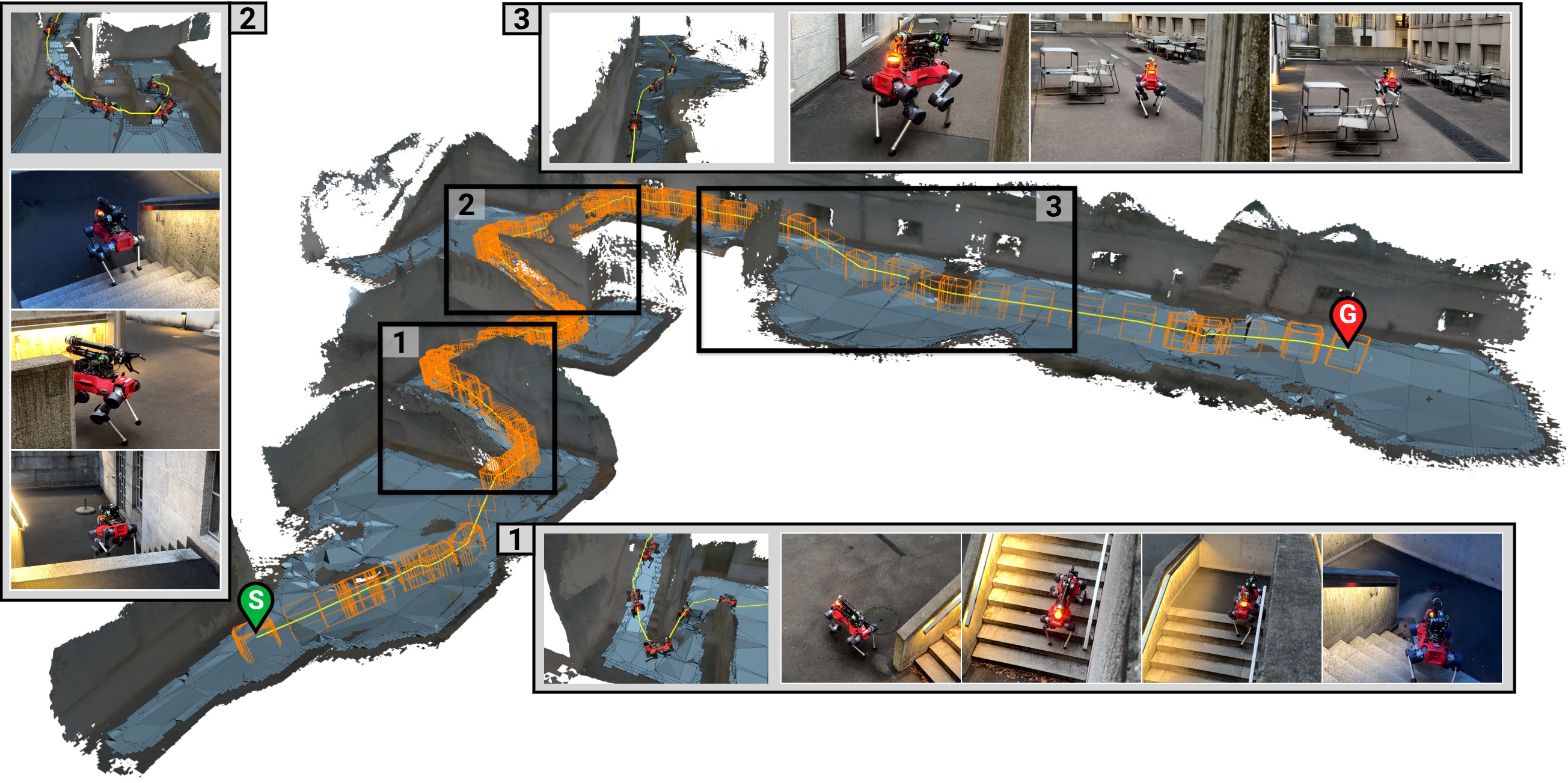}
    \caption{Real-world navigation in an outdoor multi-level environment. Starting from the lower level, the robot follows a path planned on the SE(2) NavMesh to reach the goal on the upper platform. Along the route, it ascends multiple stair segments (1 and 2) and avoids tables and chairs on the platform (3).}
    \label{fig:outdoor_task}
\end{figure*}

\begin{figure*}[!t]
    \centering
    \includegraphics[width=0.8\linewidth]{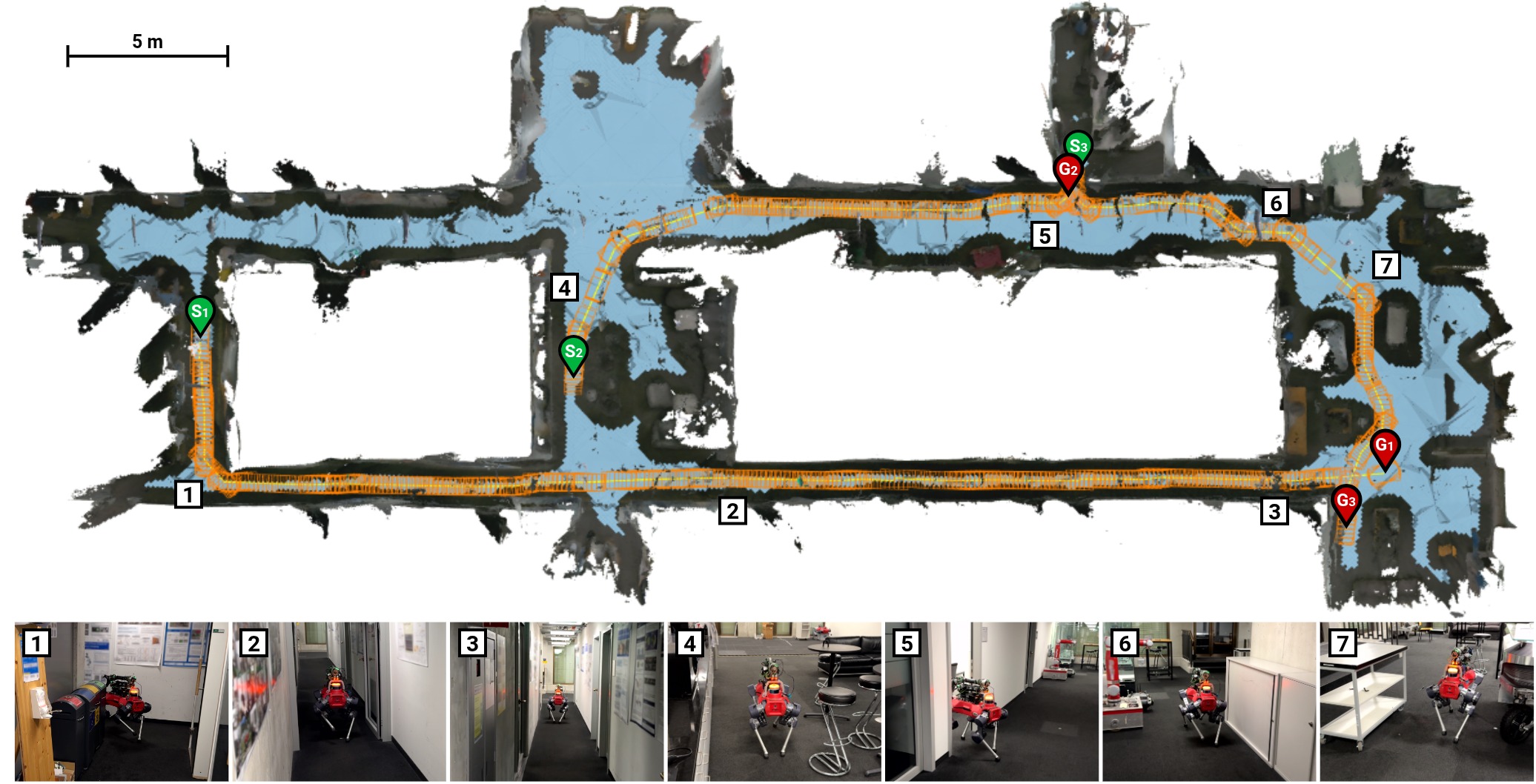}
    \caption{Real-world navigation in an indoor single-level environment. The reconstructed scene and generated SE(2) NavMesh are shown with three planned paths. The robot traverses narrow corridors (1, 3), doorways (2 and 5), and cluttered regions (4, 6, and 7).}
    \label{fig:office_task}
\end{figure*}

In the real-world experiments, we deploy the proposed online SE(2) NavMesh approach on an ANYmal platform equipped with a 6-DoF manipulator. A ZED X Mini camera is mounted on the end-effector of the manipulator to provide forward-facing perception. RGB-D images captured by the ZED X Mini are converted into colorized point clouds and downsampled to reduce computational load. The resulting point clouds are used as input to the proposed approach, with both image processing and SE(2) NavMesh generation executed onboard by an NVIDIA Jetson Orin.
Due to the additional height introduced by the manipulator, the robot height \(h_\mathrm{robot}\) is set to \SI{1.00}{\meter}. The maximum slope angle \(\theta_\mathrm{climb}\) is set to \SI{40}{\degree}, as we empirically observed that the robot can traverse steeper slopes than its nominal specifications. Voxblox parameters and slab parameters are presented in~\Cref{tab:online_gen_params}, while all remaining parameters of SE(2) NavMesh follow~\Cref{tab:sim_mesh_params}. Under this configuration, each slab aligns with four Voxblox blocks in the horizontal plane.

\begin{table}[!t]
    \centering
    \caption{Voxblox and Slab Parameters Used for Online SE(2) NavMesh Generation}
    \label{tab:online_gen_params}
    \begin{tabular}{@{} l l c @{}} 
        \toprule
        Category & Parameters & Values \\
        \midrule
        Voxblox Voxel & \( s_{\mathrm{map}} \times s_{\mathrm{map}} \times s_{\mathrm{map}} \) & \( \SI{0.05}{\meter} \times \SI{0.05}{\meter} \times \SI{0.05}{\meter} \) \\
        Voxblox Block &  \( d_{\mathrm{block}} \times d_{\mathrm{block}} \times d_{\mathrm{block}} \) & \( 16 \times 16 \times 16 \) \\
        Slab & \( d_{\mathrm{tile}} \times d_{\mathrm{tile}} \times d_{\mathrm{slab}} \) & \( 16 \times 16 \times 8 \) \\
        \bottomrule
    \end{tabular}
\end{table}

We validate the online generation capability of the proposed SE(2) NavMesh through real-world outdoor and indoor experiments.
In each experiment, the robot is first teleoperated to walk in the environment, during which the SE(2) NavMesh is incrementally generated from input point clouds. The resulting SE(2) NavMesh is then used for pathfinding to execute navigation tasks in the reconstructed environment. Instead of reconstructing it globally after each update, only the local SE(2) NavMesh for slabs affected by the latest updated geometry is generated. \Cref{fig:online_update_experiment} illustrates the process of the online SE(2) NavMesh generation in a garden.
To evaluate the computational efficiency of this local update strategy, we use the point cloud data recorded during the walk in the garden and compare the update time (defined as the total execution time from the start to the completion of each update) of local SE(2) NavMesh updates against global SE(2) NavMesh updates. \Cref{fig:online_update_comparison} shows the update time of both methods over the duration of point cloud acquisition, with the target update frequency set to \SI{4}{\hertz}. The SE(2) NavMesh update time includes geometry loading, which comprises extraction of the required triangle mesh from Voxblox and construction of the bounding volume hierarchy, as well as NavMesh construction. For both local and global methods, geometry loading dominates the update time at each update step. For global updates, the update time increases approximately linearly as time evolves, leading to a continuous growth in update latency. Once the explored map becomes sufficiently large (after approximately \SI{100}{\second} in this experiment), the system can no longer sustain the target update frequency of \SI{4}{\hertz}. In contrast, the local update method achieves a peak update time of \SI{237}{\milli\second} and an average update time of \SI{83}{\milli\second}, maintaining the \SI{4}{\hertz} update frequency throughout the entire online SE(2) NavMesh generation. The computational behavior of local updates correlates with the spatial extent of geometry changes. When the robot traverses narrow corridors, the camera observes only nearby surfaces, resulting in limited geometry updates and short local update times. In open areas, a larger portion of the environment is reconstructed at each step, increasing the number of affected slabs and the corresponding update time. These results demonstrate that the proposed online SE(2) NavMesh pipeline achieves real-time generation by restricting updates to local regions.

\Cref{fig:outdoor_task} shows the generated SE(2) NavMesh in an outdoor multi-level environment. The robot is first controlled to walk from the upper level to the lower level while incrementally generating the SE(2) NavMesh online.
Using the generated SE(2) NavMesh, a navigation task is then defined with the start state located on the lower level and the goal state on the upper platform. The planned state sequence is tracked as navigation waypoints, enabling the robot to successfully ascend multiple stair segments and return to the upper platform. \Cref{fig:office_task} presents a single-level environment containing narrow corridors, doorways, and cluttered objects. Using the generated SE(2) NavMesh, the robot successfully completes a \SI{41}{\meter} navigation task through a long narrow corridor following the planned path. In addition, the robot consecutively executes two navigation tasks that require entering and exiting the designated room through narrow doorways, with path lengths of \SI{19}{\meter} and \SI{18}{\meter}, respectively, without failure. We further evaluate the proposed approach in two challenging scenarios shown in~\Cref{fig:se2_navmesh}a and~\Cref{fig:se2_navmesh}b. The first scenario features a passage with a width of only \SI{0.8}{\meter}, requiring accurate yaw-aware navigation through tight lateral clearance. The second scenario requires the robot to navigate beneath an overhanging crane, demonstrating height-aware traversability estimation in the presence of limited overhead clearance. In both scenarios, the proposed approach accurately captures the traversable regions and generates feasible paths for the robot.

These experiments demonstrate that the proposed online SE(2) NavMesh pipeline supports real-time generation using only onboard sensing and computation. The approach is validated across diverse real-world scenes, including multi-level environments, cluttered spaces, overhangs, and narrow passages. By encoding yaw-dependent traversability, the generated SE(2) NavMesh enables reliable path planning in regions where feasible traversal strongly depends on robot heading.

\section{Limitations and Future Work}
\label{sec:limitations}
While our approach demonstrates strong performance in both simulation and real-world experiments, there are still some limitations.
Some ground robots can adjust their body height, whereas our method considers only a fixed agent height. As a result, regions that could be traversed by lowering the agent height are filtered out during traversability estimation, leading to an underestimation of the traversable area in the environment.
The current SE(2) NavMesh generation process relies solely on geometric information about the environment. Therefore, low-height objects placed on the ground may be confused with the ground surface and incorrectly classified as traversable. Incorporating semantic segmentation would help distinguish terrain from objects and differentiate terrain types. 
The proposed framework supports online SE(2) NavMesh generation from streaming sensor observations. Future work could integrate autonomous exploration strategies with the online generation process to enable fully autonomous mesh construction and navigation in previously unknown environments.

\section{Conclusions}
\label{sec:conclusion}

In this work, we presented SE(2) Navigation Mesh, a representation for global navigation of ground robots in complex 3D environments. By incorporating the robot's yaw into navigation mesh construction, SE(2) NavMesh captures yaw-dependent traversability that classical NavMeshes cannot represent. We introduced a yaw-specific layered representation that describes traversable regions under different robot headings and defines their inter-layer and intra-layer connectivity. We developed an ASA pathfinding strategy for SE(2) NavMesh to produce geometrically efficient paths while respecting yaw feasibility constraints. In addition, we proposed an online generation pipeline that constructs SE(2) NavMesh efficiently from point cloud observations. Experiments in multiple simulation scenes demonstrate that SE(2) NavMesh significantly improves traversable area coverage and connectivity compared with classical NavMeshes, while introducing only modest computational overhead. Real-world experiments further validate that SE(2) NavMesh can be generated online and can support reliable navigation of legged robots in complex environments.

\bibliographystyle{IEEEtran}
\bibliography{bibliography/references}

@article{lattanzi2017review,
  title={Review of robotic infrastructure inspection systems},
  author={Lattanzi, David and Miller, Gregory},
  journal={Journal of Infrastructure Systems},
  volume={23},
  number={3},
  pages={04017004},
  year={2017},
  publisher={American Society of Civil Engineers}
}

@article{halder2023robots,
  title={Robots in inspection and monitoring of buildings and infrastructure: A systematic review},
  author={Halder, Srijeet and Afsari, Kereshmeh},
  journal={Applied Sciences},
  volume={13},
  number={4},
  pages={2304},
  year={2023},
  publisher={MDPI}
}

@article{miura2020plant,
  title={Plant inspection by using a ground vehicle and an aerial robot: lessons learned from plant disaster prevention challenge in world robot summit 2018},
  author={Miura, Hiroyasu and Watanabe, Ayaka and Okugawa, Masayuki and Miura, Takahiko and Koganeya, Tomohiko},
  journal={Advanced Robotics},
  volume={34},
  number={2},
  pages={104--118},
  year={2020},
  publisher={Taylor \& Francis}
}

@article{kim2019uav,
  title={UAV-assisted autonomous mobile robot navigation for as-is 3D data collection and registration in cluttered environments},
  author={Kim, Pileun and Park, Jisoo and Cho, Yong K and Kang, Junsuk},
  journal={Automation in Construction},
  volume={106},
  pages={102918},
  year={2019},
  publisher={Elsevier}
}

@article{fischer2024evaluation,
  title={Evaluation of a smart mobile robotic system for industrial plant inspection and supervision},
  author={Fischer, Georg KJ and Bergau, Max and G{\'o}mez-Rosal, D Adriana and Wachaja, Andreas and Graeter, Johannes and Odenweller, Matthias and Piechottka, Uwe and H{\"o}flinger, Fabian and Gosala, Nikhil and Wetzel, Niklas and others},
  journal={IEEE Sensors Journal},
  volume={24},
  number={12},
  pages={19684--19697},
  year={2024},
  publisher={IEEE}
}

@article{miki2022wild,
    author = {Takahiro Miki  and Joonho Lee  and Jemin Hwangbo  and Lorenz Wellhausen  and Vladlen Koltun  and Marco Hutter },
    title = {Learning robust perceptive locomotion for quadrupedal robots in the wild},
    journal = {Science Robotics},
    volume = {7},
    number = {62},
    pages = {eabk2822},
    year = {2022},
    doi = {10.1126/scirobotics.abk2822}
}

@inproceedings{agarwal2023legged,
  title={Legged locomotion in challenging terrains using egocentric vision},
  author={Agarwal, Ananye and Kumar, Ashish and Malik, Jitendra and Pathak, Deepak},
  booktitle={Conference on robot learning},
  pages={403--415},
  year={2023},
  organization={PMLR}
}

@article{kolvenbach2020towards,
  title={Towards autonomous inspection of concrete deterioration in sewers with legged robots},
  author={Kolvenbach, Hendrik and Wisth, David and Buchanan, Russell and Valsecchi, Giorgio and Grandia, Ruben and Fallon, Maurice and Hutter, Marco},
  journal={Journal of field robotics},
  volume={37},
  number={8},
  pages={1314--1327},
  year={2020},
  publisher={Wiley Online Library}
}

@article{afsari2021fundamentals,
  title={Fundamentals and prospects of four-legged robot application in construction progress monitoring},
  author={Afsari, Kereshmeh and Halder, Srijeet and Ensafi, Mahnaz and DeVito, Stephen and Serdakowski, John},
  journal={EPiC series in built environment},
  volume={2},
  pages={274--283},
  year={2021}
}

@article{liu2015robotic,
  title={Robotic online path planning on point cloud},
  author={Liu, Ming},
  journal={IEEE transactions on cybernetics},
  volume={46},
  number={5},
  pages={1217--1228},
  year={2015},
  publisher={IEEE}
}

@article{krusi2017driving,
  title={{Driving on point clouds: Motion planning, trajectory optimization, and terrain assessment in generic nonplanar environments}},
  author={Kr{\"u}si, Philipp and Furgale, Paul and Bosse, Michael and Siegwart, Roland},
  journal={Journal of Field Robotics},
  volume={34},
  number={5},
  pages={940--984},
  year={2017},
  publisher={Wiley Online Library}
}

@inproceedings{ruetz2019ovpc,
  title={Ovpc mesh: 3d free-space representation for local ground vehicle navigation},
  author={Ruetz, Fabio and Hern{\'a}ndez, Emili and Pfeiffer, Mark and Oleynikova, Helen and Cox, Mark and Lowe, Thomas and Borges, Paulo},
  booktitle={2019 International Conference on Robotics and Automation (ICRA)},
  pages={8648--8654},
  year={2019},
  organization={IEEE}
}

@INPROCEEDINGS{puetz2021trianglemesh,
  author={Pütz, Sebastian and Wiemann, Thomas and Piening, Malte Kleine and Hertzberg, Joachim},
  booktitle={2021 IEEE International Conference on Robotics and Automation (ICRA)}, 
  title={Continuous Shortest Path Vector Field Navigation on 3D Triangular Meshes for Mobile Robots}, 
  year={2021},
  volume={},
  number={},
  pages={2256-2263},
  doi={10.1109/ICRA48506.2021.9560981}
}

@inproceedings{oleynikova2017voxblox,
  title={Voxblox: Incremental 3d euclidean signed distance fields for on-board mav planning},
  author={Oleynikova, Helen and Taylor, Zachary and Fehr, Marius and Siegwart, Roland and Nieto, Juan},
  booktitle={2017 IEEE/RSJ International Conference on Intelligent Robots and Systems (IROS)},
  pages={1366--1373},
  year={2017},
  organization={IEEE}
}

@article{hornung2013octomap,
  title={OctoMap: An efficient probabilistic 3D mapping framework based on octrees},
  author={Hornung, Armin and Wurm, Kai M and Bennewitz, Maren and Stachniss, Cyrill and Burgard, Wolfram},
  journal={Autonomous robots},
  volume={34},
  number={3},
  pages={189--206},
  year={2013},
  publisher={Springer}
}

@article{zhou2020ego,
  title={Ego-planner: An esdf-free gradient-based local planner for quadrotors},
  author={Zhou, Xin and Wang, Zhepei and Ye, Hongkai and Xu, Chao and Gao, Fei},
  journal={IEEE Robotics and Automation Letters},
  volume={6},
  number={2},
  pages={478--485},
  year={2020},
  publisher={IEEE}
}

@inproceedings{ren2024rog,
  title={Rog-map: An efficient robocentric occupancy grid map for large-scene and high-resolution lidar-based motion planning},
  author={Ren, Yunfan and Cai, Yixi and Zhu, Fangcheng and Liang, Siqi and Zhang, Fu},
  booktitle={2024 IEEE/RSJ International Conference on Intelligent Robots and Systems (IROS)},
  pages={8119--8125},
  year={2024},
  organization={IEEE}
}

@inproceedings{frey2022locomotion,
  title={{Locomotion policy guided traversability learning using volumetric representations of complex environments}},
  author={Frey, Jonas and Hoeller, David and Khattak, Shehryar and Hutter, Marco},
  booktitle={2022 IEEE/RSJ International Conference on Intelligent Robots and Systems (IROS)},
  pages={5722--5729},
  year={2022},
  organization={IEEE}
}

@article{chen2023smug,
  title={SMUG Planner: A Safe Multi-Goal Planner for Mobile Robots in Challenging Environments},
  author={Chen, Changan and Frey, Jonas and Arm, Philip and Hutter, Marco},
  journal={IEEE Robotics and Automation Letters},
  volume={8},
  number={11},
  pages={7170--7177},
  year={2023},
  publisher={IEEE}
}

@article{li2025real,
  title={Real-time multi-level terrain-aware path planning for ground mobile robots in large-scale rough terrains},
  author={Li, Yuxiang and Chen, Kun and Wang, Yifei and Zhang, Weifan and Wang, Jiancheng and Chen, Haoyao and Liu, Yunhui},
  journal={IEEE Transactions on Robotics},
  year={2025},
  publisher={IEEE}
}

@article{ahtiainen2017normal,
  title={Normal distributions transform traversability maps: lidar-only approach for traversability mapping in outdoor environments},
  author={Ahtiainen, Juhana and Stoyanov, Todor and Saarinen, Jari},
  journal={Journal of Field Robotics},
  volume={34},
  number={3},
  pages={600--621},
  year={2017},
  publisher={Wiley Online Library}
}

@article{elfes2002using,
  title={Using occupancy grids for mobile robot perception and navigation},
  author={Elfes, Alberto},
  journal={Computer},
  volume={22},
  number={6},
  pages={46--57},
  year={2002},
  publisher={IEEE}
}

@incollection{fankhauser2014robot,
  title={{Robot-centric elevation mapping with uncertainty estimates}},
  author={Fankhauser, P{\'e}ter and Bloesch, Michael and Gehring, Christian and Hutter, Marco and Siegwart, Roland},
  booktitle={Mobile Service Robotics},
  pages={433--440},
  year={2014},
  publisher={World Scientific}
}

@article{fankhauser2018probabilistic,
  title={Probabilistic terrain mapping for mobile robots with uncertain localization},
  author={Fankhauser, P{\'e}ter and Bloesch, Michael and Hutter, Marco},
  journal={IEEE Robotics and Automation Letters},
  volume={3},
  number={4},
  pages={3019--3026},
  year={2018},
  publisher={IEEE}
}

@inproceedings{miki2022elevation,
  title={{Elevation mapping for locomotion and navigation using gpu}},
  author={Miki, Takahiro and Wellhausen, Lorenz and Grandia, Ruben and Jenelten, Fabian and Homberger, Timon and Hutter, Marco},
  booktitle={2022 IEEE/RSJ International Conference on Intelligent Robots and Systems (IROS)},
  pages={2273--2280},
  year={2022},
  organization={IEEE}
}

@ARTICLE{Yang2025tomography,
  author={Yang, Bowen and Cheng, Jie and Xue, Bohuan and Jiao, Jianhao and Liu, Ming},
  journal={IEEE/ASME Transactions on Mechatronics}, 
  title={Efficient Global Navigational Planning in 3-D Structures Based on Point Cloud Tomography}, 
  year={2025},
  volume={30},
  number={1},
  pages={321-332},
  doi={10.1109/TMECH.2024.3396001}
}

@misc{unity,
  author = {{Unity Technologies}},
  title = {Navigation Overview},
  howpublished = {\url{https://docs.unity3d.com/Packages/com.unity.ai.navigation@2.0/manual/NavigationOverview.html}},
  year = {2026}
}

@misc{unreal,
  author = {{Epic Games}},
  title = {Navigation System},
  howpublished = {\url{https://dev.epicgames.com/documentation/en-us/unreal-engine/navigation-system-in-unreal-engine}},
  year = {2024}
}

@misc{habitatchallenge2023,
  title = {Habitat Challenge 2023},
  author = {Karmesh Yadav and Jacob Krantz and Ram Ramrakhya and Santhosh Kumar Ramakrishnan and Jimmy Yang and Austin Wang and John Turner and Aaron Gokaslan and Vincent-Pierre Berges and Roozbeh Mootaghi and Oleksandr Maksymets and Angel X Chang and Manolis Savva and Alexander Clegg and Devendra Singh Chaplot and Dhruv Batra},
  howpublished = {\url{https://aihabitat.org/challenge/2023/}},
  year = {2023}
}

@inproceedings{van2016comparative,
  title={{A comparative study of navigation meshes}},
  author={Van Toll, Wouter and Triesscheijn, Roy and Kallmann, Marcelo and Oliva, Ramon and Pelechano, Nuria and Pettr{\'e}, Julien and Geraerts, Roland},
  booktitle={Proceedings of the 9th International Conference on Motion in Games},
  pages={91--100},
  year={2016}
}

@article{hart1968formal,
  title={A formal basis for the heuristic determination of minimum cost paths},
  author={Hart, Peter E and Nilsson, Nils J and Raphael, Bertram},
  journal={IEEE transactions on Systems Science and Cybernetics},
  volume={4},
  number={2},
  pages={100--107},
  year={1968},
  publisher={IEEE}
}

@article{ramakrishnan2021habitat,
  title={Habitat-matterport 3d dataset (hm3d): 1000 large-scale 3d environments for embodied ai},
  author={Ramakrishnan, Santhosh K and Gokaslan, Aaron and Wijmans, Erik and Maksymets, Oleksandr and Clegg, Alex and Turner, John and Undersander, Eric and Galuba, Wojciech and Westbury, Andrew and Chang, Angel X and others},
  journal={arXiv preprint arXiv:2109.08238},
  year={2021}
}

@article{khanna2023hssd,
  author={{Khanna*}, Mukul and {Mao*}, Yongsen and Jiang, Hanxiao and Haresh, Sanjay and Shacklett, Brennan and Batra, Dhruv and Clegg, Alexander and Undersander, Eric and Chang, Angel X. and Savva, Manolis},
  title={{Habitat Synthetic Scenes Dataset (HSSD-200): An Analysis of 3D Scene Scale and Realism Tradeoffs for ObjectGoal Navigation}},
  journal={arXiv preprint},
  year={2023},
  eprint={2306.11290},
  archivePrefix={arXiv},
  primaryClass={cs.CV}
}

@article{brandao2020gaitmesh,
  title={{GaitMesh: Controller-aware navigation meshes for long-range legged locomotion planning in multi-layered environments}},
  author={Brandao, Martim and Aladag, Omer Burak and Havoutis, Ioannis},
  journal={IEEE Robotics and Automation Letters},
  volume={5},
  number={2},
  pages={3596--3603},
  year={2020},
  publisher={IEEE}
}

@misc{mononen2014recastnav,
  author = {Mononen, Mikko},
  title = {{Recast Navigation}},
  howpublished = {\url{https://github.com/recastnavigation/recastnavigation}},
  year = {2014}    
}

@article{cui2011based,
  title={A*-based pathfinding in modern computer games},
  author={Cui, Xiao and Shi, Hao},
  journal={International Journal of Computer Science and Network Security},
  volume={11},
  number={1},
  pages={125--130},
  year={2011},
  publisher={International Journal of Computer Science and Network Security (IJCSNS)}
}

@article{perez20193d,
  title={3D exploration and navigation with optimal-RRT planners for ground robots in indoor incidents},
  author={P{\'e}rez-Higueras, No{\'e} and Jard{\'o}n, Alberto and Rodr{\'\i}guez, {\'A}ngel and Balaguer, Carlos},
  journal={Sensors},
  volume={20},
  number={1},
  pages={220},
  year={2019},
  publisher={MDPI}
}

@inproceedings{yang2022far,
  title={Far planner: Fast, attemptable route planner using dynamic visibility update},
  author={Yang, Fan and Cao, Chao and Zhu, Hongbiao and Oh, Jean and Zhang, Ji},
  booktitle={2022 ieee/rsj international conference on intelligent robots and systems (iros)},
  pages={9--16},
  year={2022},
  organization={IEEE}
}

@article{kong2023marsim,
  title={MARSIM: A light-weight point-realistic simulator for LiDAR-based UAVs},
  author={Kong, Fanze and Liu, Xiyuan and Tang, Benxu and Lin, Jiarong and Ren, Yunfan and Cai, Yixi and Zhu, Fangcheng and Chen, Nan and Zhang, Fu},
  journal={IEEE Robotics and Automation Letters},
  volume={8},
  number={5},
  pages={2954--2961},
  year={2023},
  publisher={IEEE}
}

@article{wand2008processing,
  title={Processing and interactive editing of huge point clouds from 3D scanners},
  author={Wand, Michael and Berner, Alexander and Bokeloh, Martin and Jenke, Philipp and Fleck, Arno and Hoffmann, Mark and Maier, Benjamin and Staneker, Dirk and Schilling, Andreas and Seidel, Hans-Peter},
  journal={Computers \& Graphics},
  volume={32},
  number={2},
  pages={204--220},
  year={2008},
  publisher={Elsevier}
}

@article{tuna2023x,
  title={X-icp: Localizability-aware lidar registration for robust localization in extreme environments},
  author={Tuna, Turcan and Nubert, Julian and Nava, Yoshua and Khattak, Shehryar and Hutter, Marco},
  journal={IEEE Transactions on Robotics},
  volume={40},
  pages={452--471},
  year={2023},
  publisher={IEEE}
}

@inproceedings{jelavic2022open3d,
  title={Open3d slam: Point cloud based mapping and localization for education},
  author={Jelavic, Edo and Nubert, Julian and Hutter, Marco},
  booktitle={Robotic Perception and Mapping: Emerging Techniques, ICRA 2022 Workshop},
  pages={24},
  year={2022},
  organization={ETH Zurich, Robotic Systems Lab}
}

@inproceedings{medioni2000tensor,
  title={Tensor voting: Theory and applications},
  author={Medioni, G{\'e}rard and Tang, Chi-Keung and Lee, Mi-Suen},
  booktitle={Proceedings of RFIA},
  volume={2000},
  year={2000},
  organization={Hermes, Lavoisier Paris, France}
}

@article{dijkstra1959note,
  title={A note on two problems in connexion with graphs},
  author={Dijkstra, EW},
  journal={Numerische Mathematik},
  volume={1},
  number={1},
  pages={269--271},
  year={1959}
}

@article{lavalle2001rapidly,
  title={Rapidly-Exploring Random Trees: Progress and Prospects},
  author={LaValle, Steven M},
  journal={Algorithmic and Computational Robotics: New Directions 2000 WAFR},
  pages={293},
  year={2001},
  publisher={CRC Press}
}

@article{karaman2011sampling,
  title={Sampling-based algorithms for optimal motion planning},
  author={Karaman, Sertac and Frazzoli, Emilio},
  journal={The international journal of robotics research},
  volume={30},
  number={7},
  pages={846--894},
  year={2011},
  publisher={Sage Publications Sage UK: London, England}
}

@inproceedings{hoppe1992surface,
  title={Surface reconstruction from unorganized points},
  author={Hoppe, Hugues and DeRose, Tony and Duchamp, Tom and McDonald, John and Stuetzle, Werner},
  booktitle={Proceedings of the 19th annual conference on computer graphics and interactive techniques},
  pages={71--78},
  year={1992}
}

@inproceedings{wiemann2018surface,
  title={Surface reconstruction from arbitrarily large point clouds},
  author={Wiemann, Thomas and Mitschke, Isaak and Mock, Alexander and Hertzberg, Joachim},
  booktitle={2018 Second IEEE International Conference on Robotic Computing (IRC)},
  pages={278--281},
  year={2018},
  organization={IEEE}
}

@inproceedings{kuffner2000rrt,
  title={RRT-connect: An efficient approach to single-query path planning},
  author={Kuffner, James J and LaValle, Steven M},
  booktitle={Proceedings 2000 ICRA. Millennium conference. IEEE international conference on robotics and automation. Symposia proceedings (Cat. No. 00CH37065)},
  volume={2},
  pages={995--1001},
  year={2000},
  organization={IEEE}
}

@article{niessner2013real,
  title={Real-time 3D reconstruction at scale using voxel hashing},
  author={Nie{\ss}ner, Matthias and Zollh{\"o}fer, Michael and Izadi, Shahram and Stamminger, Marc},
  journal={ACM Transactions on Graphics (ToG)},
  volume={32},
  number={6},
  pages={1--11},
  year={2013},
  publisher={ACM New York, NY, USA}
}

@article{duberg2020ufomap,
  title={UFOMap: An efficient probabilistic 3D mapping framework that embraces the unknown},
  author={Duberg, Daniel and Jensfelt, Patric},
  journal={IEEE Robotics and Automation Letters},
  volume={5},
  number={4},
  pages={6411--6418},
  year={2020},
  publisher={IEEE}
}

@article{tang2026memory,
  title={Memory-efficient boundary map for large-scale occupancy grid mapping},
  author={Tang, Benxu and Ren, Yunfan and Cai, Yixi and Kong, Fanze and Liu, Wenyi and Zhu, Fangcheng and Yin, Longji and Shi, Liuyu and Zhang, Fu},
  journal={The International Journal of Robotics Research},
  pages={02783649261425266},
  year={2026},
  publisher={SAGE Publications Sage UK: London, England}
}

@inproceedings{wang2023towards,
  title={Towards Efficient Trajectory Generation for Ground Robots beyond 2D Environment},
  author={Wang, Jingping and Xu, Long and Fu, Haoran and Meng, Zehui and Xu, Chao and Cao, Yanjun and Lyu, Ximin and Gao, Fei},
  booktitle={2023 IEEE International Conference on Robotics and Automation (ICRA)},
  pages={7858--7864},
  year={2023},
  organization={IEEE}
}

@inproceedings{lu2014layered,
  title={Layered costmaps for context-sensitive navigation},
  author={Lu, David V and Hershberger, Dave and Smart, William D},
  booktitle={2014 IEEE/RSJ International Conference on Intelligent Robots and Systems},
  pages={709--715},
  year={2014},
  organization={IEEE}
}

@inproceedings{wellhausen2021rough,
  title={Rough terrain navigation for legged robots using reachability planning and template learning},
  author={Wellhausen, Lorenz and Hutter, Marco},
  booktitle={2021 IEEE/RSJ International Conference on Intelligent Robots and Systems (IROS)},
  pages={6914--6921},
  year={2021},
  organization={IEEE}
}

@inproceedings{hauser2015lazy,
  title={Lazy collision checking in asymptotically-optimal motion planning},
  author={Hauser, Kris},
  booktitle={2015 IEEE international conference on robotics and automation (ICRA)},
  pages={2951--2957},
  year={2015},
  organization={IEEE}
}

@inproceedings{yang2021real,
  title={Real-time optimal navigation planning using learned motion costs},
  author={Yang, Bowen and Wellhausen, Lorenz and Miki, Takahiro and Liu, Ming and Hutter, Marco},
  booktitle={2021 IEEE International Conference on Robotics and Automation (ICRA)},
  pages={9283--9289},
  year={2021},
  organization={IEEE}
}

@inproceedings{beucher1979use,
  title={Use of watersheds in contour detection},
  author={Beucher, Serge},
  booktitle={Proc. Int. Workshop on Image Processing, Sept. 1979},
  pages={17--21},
  year={1979}
}

@article{eberly2008triangulation,
  title={Triangulation by ear clipping},
  author={Eberly, David},
  journal={Geometric Tools},
  pages={2002--2005},
  year={2008}
}

@misc{mononen2010funnel,
  author={Mikko Mononen},
  title={{Simple Stupid Funnel Algorithm}},
  howpublished={\url{https://digestingduck.blogspot.com/2010/03/simple-stupid-funnel-algorithm.html}},
  year={2010}
}

@article{lorensen1987marching,
  title={Marching cubes: A high resolution 3D surface construction algorithm},
  author={Lorensen, William E and Cline, Harvey E},
  journal={ACM SIGGRAPH Computer Graphics},
  volume={21},
  number={4},
  pages={163--169},
  year={1987},
  publisher={Association for Computing Machinery (ACM)}
}

@inproceedings{hutter2016anymal,
  title={Anymal-a highly mobile and dynamic quadrupedal robot},
  author={Hutter, Marco and Gehring, Christian and Jud, Dominic and Lauber, Andreas and Bellicoso, C Dario and Tsounis, Vassilios and Hwangbo, Jemin and Bodie, Karen and Fankhauser, Peter and Bloesch, Michael and others},
  booktitle={2016 IEEE/RSJ international conference on intelligent robots and systems (IROS)},
  pages={38--44},
  year={2016},
  organization={IEEE}
}

@misc{anymal2022specs,
  author={ANYbotics},
  title={ANYmal Technical Specifications},
  howpublished = {\url{https://www.anybotics.com/anymal-technical-specifications.pdf}},
  year={2022}
}

@article{sucan2012open,
  title={The open motion planning library},
  author={Sucan, Ioan A and Moll, Mark and Kavraki, Lydia E},
  journal={IEEE Robotics \& Automation Magazine},
  volume={19},
  number={4},
  pages={72--82},
  year={2012},
  publisher={IEEE}
}

@article{kavraki2002probabilistic,
  title={Probabilistic roadmaps for path planning in high-dimensional configuration spaces},
  author={Kavraki, Lydia E and Svestka, Petr and Latombe, J-C and Overmars, Mark H},
  journal={IEEE transactions on Robotics and Automation},
  volume={12},
  number={4},
  pages={566--580},
  year={2002},
  publisher={IEEE}
}

@article{anderson2018evaluation,
  title={On evaluation of embodied navigation agents},
  author={Anderson, Peter and Chang, Angel and Chaplot, Devendra Singh and Dosovitskiy, Alexey and Gupta, Saurabh and Koltun, Vladlen and Kosecka, Jana and Malik, Jitendra and Mottaghi, Roozbeh and Savva, Manolis and others},
  journal={arXiv preprint arXiv:1807.06757},
  year={2018}
}

@inproceedings{yokoyama2021success,
  title={Success weighted by completion time: A dynamics-aware evaluation criteria for embodied navigation},
  author={Yokoyama, Naoki and Ha, Sehoon and Batra, Dhruv},
  booktitle={2021 IEEE/RSJ International Conference on Intelligent Robots and Systems (IROS)},
  pages={1562--1569},
  year={2021},
  organization={IEEE}
}

\end{document}